\documentclass[preprint,12pt]{elsarticle}

\usepackage{amsmath}
\usepackage{amssymb}
\usepackage{graphicx}
\usepackage{tikz}
\usepackage{subfig}
\usepackage{rotating}
\usepackage{multirow}
\usepackage{ogonek}
\usepackage{url}
\usepackage{float} 
\usepackage{picins}
\usetikzlibrary{arrows,positioning}
 \usepackage{color}

\tikzset{
    >=stealth',
    blackarrow/.style={
           ->,
           thick,
           shorten <=2pt,
           shorten >=2pt,}
}

\newcommand{\website}{\url{http://www.miproblems.org}}

\definecolor{myblue}{rgb}{.0,.0,.0}
\newcommand{\added}[1]{\textcolor{myblue}{#1}}

\journal{Pattern Recognition}

\begin{document}

\begin{frontmatter}



\title{Multiple Instance Learning with Bag Dissimilarities}

\address[prlab]{Pattern Recognition Laboratory, Delft University of Technology, The Netherlands}
\address[diku]{Image Group, University of Copenhagen, Denmark}

\author[prlab]{Veronika Cheplygina\corref{cor1}}
\ead{v.cheplygina@tudelft.nl}
\cortext[cor1]{Corresponding author.\\ \emph{Address:} Mekelweg 4, 2628CD Delft, The Netherlands. \emph{Phone:} +31152787243}

\author[prlab]{David M.J. Tax}
\ead{d.m.j.tax@tudelft.nl}

\author[prlab,diku]{Marco Loog}
\ead{m.loog@tudelft.nl}

\begin{abstract}
%200 word limit, now 164
Multiple instance learning (MIL) is concerned with learning from sets (bags) of objects (instances), where the individual instance labels are ambiguous. In this setting, supervised learning cannot be applied directly. Often, specialized MIL methods learn by making additional assumptions about the relationship of the bag labels and instance labels. Such assumptions may fit a particular dataset, but do not generalize to the whole range of MIL problems. Other MIL methods shift the focus of assumptions from the labels to the overall (dis)similarity of bags, and therefore learn from bags directly. We propose to represent each bag by a vector of its dissimilarities to other bags in the training set, and treat these dissimilarities as a feature representation. We show several alternatives to define a dissimilarity between bags and discuss which definitions are more suitable for particular MIL problems. The experimental results show that the proposed approach is computationally inexpensive, yet very competitive with state-of-the-art algorithms on a wide range of MIL datasets. 

\end{abstract}


\end{frontmatter}

%%%% MAX 35 pages

\section{Introduction}

%[TODO: Problems and contributions, do not repeat description of methods]

Many pattern recognition problems deal with complex objects that consist of parts: images displaying several objects, documents with different paragraphs, proteins with various amino acid subsequences. The success of supervised learning techniques forces such complex objects to be represented as a single feature vector. However, this reduction may cause important differences in objects to be lost, degrading classification performance. Rather than representing such a complex object by a single feature vector, we can represent it by a set of feature vectors, as in multiple instance, or multi-instance learning (MIL) ~\cite{dietterich1997solving}. For example, an image can be represented as a bag of segments, where each segment is represented by its own feature vector. This is a more flexible representation that potentially can preserve more information than a single feature vector representation. 

In MIL terminology, an object is called a \emph{bag} and its feature vectors are called \emph{instances}. MIL problems are often considered to be two-class problems, i.e., a bag can belong either to the positive or the negative class. During training, the bag labels are available, but the labels of the instances are unknown. Often assumptions are made about the instance labels and their relationship with the bag labels. The standard assumption is that positive bags contain at least one positive or \emph{concept} instance, whereas negative bags contain only negative, \emph{background} instances ~\cite{dietterich1997solving,maron1998framework}. An image labeled as ``tiger'' would therefore contain a tiger in at least one of its segments, whereas images with other labels would not depict any tigers. Many MIL approaches therefore focus on using the labeled bags to model the concept region in the instance space. To classify a previously unseen bag, the instances are labeled according to the best candidate model for the concept, and the bag label is then obtained from these instance labels.

It has been pointed out~\cite{chen2006miles} that although for many problems the bag representation is useful, the assumptions on the bag and instance labels typically do not fit the application. For instance, for an image of the ``desert'' category, it would be wrong to say that ``sand'' is the concept, if images of the ``beach'' category are also present. Therefore, methods in which the standard assumption is relaxed, have emerged. In ~\cite{wang2008adaptive} an adaptive parameter is used to determine the fraction of concept instances in positive bags. Generalized MIL~\cite{weidmann2003two,scott2005generalized} examines the idea that there could be an arbitrary number of concepts, where each concept has a rule of how many (just one, several or a fraction) positive instances are needed to satisfy each concept. A review of MIL assumptions can be found in ~\cite{foulds2010review}.

This line of thought can be extended further to cases where it is difficult to define a concept or concepts, and where most, if not all, instances, contribute to the bag label. The implicit assumption is that bag labels can be related to distances between bags, or to distances between bags and instances. Such approaches have used bag distances~\cite{wang2000solving}, bag kernels~\cite{gartner2002multi}, instance kernels~\cite{chen2006miles} or dissimilarities between bags~\cite{tax2011bag,cheplygina2012does,sorensen2010dissimilarity}. %This is the most general case, and also fits best to how we see MIL (learning from sets of instances, without extra assumptions about the instance labels).

Bag-based approaches are attractive because because they transform the MIL dataset back to a standard feature vector representation such that regular supervised classifiers can be used. Unfortunately, some of the representational power of MIL can be lost when converting a bag to a single feature vector of (dis)similarities. It has indeed been pointed out that the definition of distance or similarity can influence how well the representation is suited for one or more concepts~\cite{foulds2010review}. The question is how to do this in a way that still preserves information about the class differences. Furthermore, competing approaches offer a variety of definitions of (dis)similarity, and it is not always clear which definition should be preferred when a new type of MIL problem presents itself. 

In this paper we propose a general MIL dissimilarity approach called MInD (\textbf{M}ultiple \textbf{In}stance \textbf{D}issimilarity). We discuss several ways in which dissimilarities between bags can be defined, show which assumptions these definitions are implicitly making, and hence which definitions are suitable for different types of MIL problems. We have collected several examples of such problems in a single repository online\footnote{http://www.miproblems.org}.  Furthermore, we discuss why the dissimilarity space is an attractive approach for MIL in general. An important advantage is that there are no restrictions on the dissimilarity measure (such as metricity or positive-definiteness). This allows the use of expert-defined dissimilarities which often violate these mathematical restrictions. Similarly, there is no restriction on the classifier used in the dissimilarity space, which is attractive for potential end-users faced with MIL problems, and who already have experience with a certain supervised classifier. Lastly, with a suitable choice of dissimilarity and classifier, the approach is very computationally efficient, yet still provides interpretable state-of-the-art results on many datasets. For example, the average minimum distance between bags with a logistic classifier achieves very good performances, is easy to implement, and the classifier decisions can be explained in terms of dissimilarities to the prototypes. 

%Why dissimilarity is particularly so important to multi-instance learning? I still could not appreciate it even though I had read the responses. Please list 3 most important reasons to support this motivation, no matter from what aspects, for example, performance improvement, efficiency, theoretical supports, etc.

After a review of MIL approaches in Section~\ref{sec:review}, we propose MInD in Section ~\ref{sec:representation}. In Section~\ref{sec:behave}, we show some examples of MIL problems and demonstrate which dissimilarities are most suitable in each case. We then compare results to a range of MIL methods in Section ~\ref{sec:results}, and discuss practical issues of dissimilarities and other bag-level methods in Section~\ref{sec:compare}. A conclusion is given in Section~\ref{sec:conclusion}.

\section{Review of MIL Approaches}\label{sec:review}

%[TODO: More notation, instance vs bag classifier]

In multiple instance learning (MIL), an object is represented by a bag $B_i  = \{\mathbf{x}_{ik}| k=1,...,n_i\} \subset \mathbb{R}^d$ of $n_i$ feature vectors or instances. The training set $\mathcal{T} = \{(B_i, y_i) | i=1,...N\}$ consists of positive ($y_i = +1$) and negative ($y_i = -1$) bags. We will also denote such bags by $B^{+}_i$ and $B^{-}_i$. The standard assumption for MIL is that there are instance labels which relate to the bag labels as follows: a bag is positive if and only if it contains at least one positive, or \emph{concept} instance~\cite{dietterich1997solving}. 

Under this standard assumption, the strategy has been to model the concept: a region in the feature space which contains at least one instance from each positive bag, but no instances from negative bags. The original class of MIL methods used an axis-parallel hyper-rectangle (APR)~\cite{dietterich1997solving} as a model for the concept, and several search strategies involving such APRs have been proposed.

Diverse Density~\cite{maron1998framework} is another approach for finding the concept in instance space. For a given point $t$ in this space, a measure $DD(t)$ is defined as the ratio between the number of positive bags which have instances near $t$, and the distance of the negative instances to $t$. The point of maximum Diverse Density should therefore correspond to the target concept. The maximization problem does not have a closed form solution and gradient ascent is used to find the maximum. The search may therefore converge to a local optimum, and several restarts are needed to find the best solution. 

EM-DD~\cite{zhang2001dd} is an expectation-maximization algorithm that refines Diverse Density. The instance labels are modeled by hidden variables. After an initial guess for the concept $t$, the expectation step selects the most positive instance from each bag according to $t$. The maximization step then finds a new concept $t'$ by maximizing DD on the selected, most positive instances. The steps are repeated until the algorithm converges. %EM-DD has shown to perform well on a range of MIL problems, but is also computationally intensive. 

Furthermore, several regular supervised classifiers have been extended to work in the MIL setting. One example is mi-SVM~\cite{andrews2002support}, an extension of support vector machines which attempts to find hidden labels of the instances under constraints posed by the bag labels. Another example is MILBoost~\cite{viola2006multiple}, where the instances are reweighted in each of the boosting rounds. The bag labels are decided by applying a noisy OR~\cite{maron1998framework} rule to the instance labels, which reflects the standard assumption. 
%MILboost is meant for object detection, so the 1 instance only makes sense

It has been recognized that the standard assumption might be too strict for certain types of MIL problems. Therefore, relaxed assumptions have emerged~\cite{weidmann2003two,scott2005generalized}, where a fraction or a particular number of positive instances are needed to satisfy a concept, and where multiple concept regions are considered.

A similar notion is used in MILES~\cite{chen2006miles}, where multiple concepts, as well as so-called negative concepts (concepts that only negative bags have) are allowed. All of the instances in the training set are used as candidate concept targets, and each bag is represented by its similarities to these instances. A sparse 1-norm SVM is then used to simultaneously maximize the bag margin, and select the most discriminative similarities, i.e., instances that are identified as positive or negative concepts. 

A step further are methods that do not make explicit assumptions about the instances or the concepts, but only assume that bags of the same class are somehow similar to each other, and then learn from distances or similarities between bags. Such methods include Citation-$k$NN~\cite{wang2000solving}, which is based on the Hausdorff distance between bags, bag kernels~\cite{gartner2002multi} and bag dissimilarities~\cite{tax2011bag,zhang2009multi2}. In ~\cite{gartner2002multi}, a bag kernel is defined either as a sum of the instance kernels, or as a standard (linear or RBF kernel) on a transformed, single instance representation of the bag. One example is the Minimax representation, where each bag is represented by the minimum and maximum feature values of its instances. 

Last but not least, a way to learn in MIL problems is to propagate the bag labels to the instances, and use supervised learners on these propagated labels. We call this approach SimpleMIL. To obtain a bag label from predicted instance labels, the instance labels have to be combined. Here, the noisy OR rule or other combining methods can be used~\cite{li2012multiple,loog2004static}. It has been demonstrated that supervised methods can be quite effective in dealing with MIL problems~\cite{ray2005supervised}. 

\begin{figure}[ht]
 \centering

\subfloat[Original MIL]{
  \includegraphics{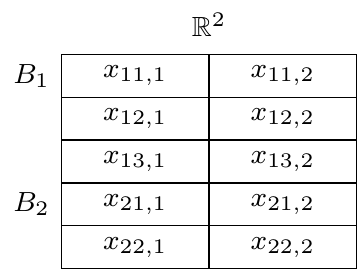}
}
\subfloat[Bag dissimilarity]{
  \includegraphics{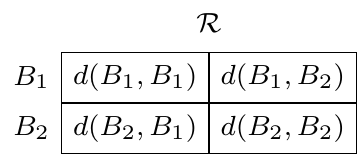}}

\subfloat[Instance dissimilarity]{
	  \includegraphics{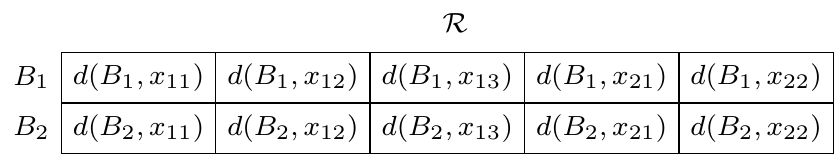}
}
\caption[]{Representations of a MIL problem with 2 bags and 2 features. $B_1$ has 3 instances and $B_2$ has 2 instances. The dimensionality of the original representation depends on the number of features, while in the dissimilarity representation, the dimensionality depends on the number of bags or instances.}
\label{fig:rep}
\end{figure}

All MIL methods can be more globally summarized by the representation that they use: the standard instance-vector-based representation, a bag dissimilairity representation and a bag-instance representation (see Fig.~\ref{fig:rep}). The first representation is the standard representation of MIL, where each bag consists of several instances, and the dimensionality is equal to the dimensionality of the instance space. In this example, there are two bags which are represented in a 2D feature space. This is the representation used by EM-DD, mi-SVM, MILBoost and SimpleMIL. The representation on the top right is the bag representation, used by Citation-$k$NN, bag kernels and our bag dissimilarity approach. The representation in the bottom is the instance representation, used by MILES. In the latter two representations, regular supervised classifiers are again applicable. In these cases, less assumptions about the relationships of bag and instance labels are needed, but the definition of (dis)similarity creates implicit assumptions on which instances are important.

\section{Bag Dissimilarity Representation}\label{sec:representation}

We can represent an object, and therefore also a MIL bag $B_i$, by its dissimilarities to prototype objects in a representation set $\mathcal{R}$~\cite{pekalska2005dissimilarity}. In our work, $\mathcal{R}$ is taken to be a subset of size $M$ of the training set $\mathcal{T}$ of size $N$ (typically $M \leq N$).  Using these $M$ prototypes, each bag is represented as $\mathbf{d}(B_i, \mathcal{T})= [d(B_i, B_1), ... d(B_i, B_M)]$: a vector of $M$ dissimilarities. Therefore, each bag is represented by a single feature vector and the MIL problem can be viewed as a regular supervised learning problem.

\added{We propose a framework which encompasses different ways to define $d(B_i,B_j)$, the dissimilarity between two bags $B_i$ and $B_j$. There are two main steps: the representation of a bag, and given this representation, the definition of the dissimilarity function. We distinguish the following approaches to treat the bags:}

\begin{itemize}

\item \added{As a point set, or subset of the feature space. In this case, $d$ is defined through a set distance.}

\item \added{As a distribution of instances. Here, $d$ is defined through a distribution distance.} 

\item \added{As an attributed graph, where the instances are the nodes, and relationships between instances are the edges~\cite{zhou2009multi}. In this case, $d$ is defined as a graph kernel or distance. However, because it is not straightforward to define the edges and determine the trade-off of nodes and edges in such a problem~\cite{lee2012bridging}, we do not focus on this case here.} 

\end{itemize}

%The problem is how to define $d(B_i,B_j)$, the dissimilarity between two bags $B_i$ and $B_j$. There are several approaches to do this. The first approach is to see each bag as a point set or a subset of the feature space. Alternatively, a bag can be seen as a distribution of instances, and distribution distances can be used to define $d(B_i,B_j)$. \added{}

\subsection{Bags as Point Sets}\label{sec:pointsets}

The first approach to define a dissimilarity of two bags is to consider each bag as a point set or a subset of a high-dimensional space. \added{One possible distance that can be computed is based on the Hausdorff metric, under which two point sets $B_i$ and $B_j$ are close to each other when every point in $B_i$ is close to a point in $B_j$}. Closeness is defined through the distance $d$ employed, which typically is Euclidean. The Hausdorff distance, derived from the metric, has been widely used in object matching in computer vision~\cite{huttenlocher1993comparing,dubuisson1994modified}. The Hausdorff distance applied to bags uses the maximum mismatch between the instances of the respective bags:

\begin{equation}
d_{h}(B_i, B_j)  =  \max_{k} \min_{l} d(\mathbf{x}_{ik}, \mathbf{x}_{jl}).
\end{equation}

As $d_h$ is not symmetric, the final Hausdorff distance is symmetrized by taking the maximum of the directed distances. All of these steps ensure that the Hausdorff distance is metric, i.e., that it satisfies the identity, symmetry and triangle inequality properties.

%[Which step is responsible for which property?] 
It has been pointed out~\cite{dubuisson1994modified} that the maximum operation in the original definition might be too sensitive to outliers, and modified versions of the Hausdorff distance have been introduced~\cite{dubuisson1994modified,zhao2005new}. Two alternative examples of defining $d(B_i,B_j)$ are shown in (\ref{eq:minmin}) and (\ref{eq:meanmin}). \added{The underlying instance dissimilarity function we use here is the squared Euclidean distance:}

\begin{equation}
d_{minmin} (B_i, B_j)  =  \min_{k}\min_{l}	d(\mathbf{x}_{ik}, \mathbf{x}_{jl})
\label{eq:minmin}
\end{equation}

and

\begin{equation}
d_{meanmin}(B_i, B_j)  =  \frac{1}{n_i}\sum_{k=1}^{n_i} \min_{l} d(\mathbf{x}_{ik}, \mathbf{x}_{jl}). 
\label{eq:meanmin}
\end{equation}

\begin{figure}[ht]
\centering
	
	\subfloat{
	  \includegraphics{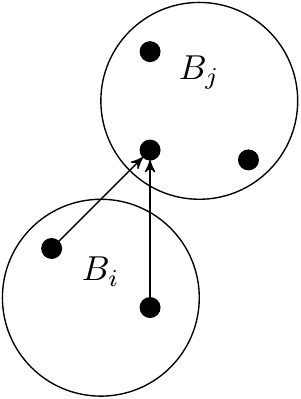}
}
\subfloat{
	  \includegraphics{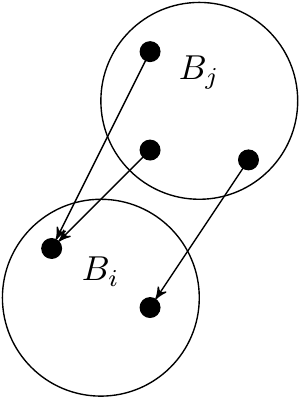}
}

  \caption{Minimum instance distances between two bags. The bag dissimilarity
is defined as the minimum, maximum, average or other statistic of these
distances. The directions of the arrows show that there are two, possibly asymmetric,
dissimilarities: that of $B_i$ to $B_j$, and that of $B_j$ to $B_i$}
\label{fig:bagdist}
 \end{figure}

Figure ~\ref{fig:bagdist} shows the first step in computing such dissimilarities between two bags. The arrows in each diagram are the minimum distances between instances of two bags, which are asymmetric. For example, in the left diagram, the two instances in $B_i$ have the same nearest instance in bag $B_j$, but this is not true for the diagram on the right. If the next step is to take the minimum of these instance distances, as in (\ref{eq:minmin}), the resulting dissimilarity will be symmetric. However, $d_{minmin}$ is symmetric, it does not satisfy the identity property, that is, $d_{minmin}(B_i,B_j) = 0$ does not imply $B_i=B_j$ when an instance in $B_i$ coincides with an instance in $B_j$, and the triangle inequality is not always satisfied. On the other hand, if we average the minimum distances as in (\ref{eq:meanmin}), we will notice that $d_{meanmin}(B_i, B_j)\neq d_{meanmin}(B_j, B_i)$.

These dissimilarities are therefore non-metric, which can be problematic when classifying bags based on the labels of their nearest neighbors. These non-metric properties are not a problem in the dissimilarity approach, because the dissimilarity matrix is viewed as a collection of features and the relation between features is not constrained like is for distances. In fact, because we step away from the requirements of a distance, we can define even more general ways to obtain $d(B_i,B_j)$ from the pairwise instance dissimilarities. 

First of all, there is no need to symmetrize the dissimilarities. In fact, it can be the case that one of the directions, i.e., measuring to the prototypes, or measuring from the prototypes, is more informative~\cite{cheplygina2012class}. If both directions are informative, this information can be combined in more ways than by just enforcing symmetry. For instance, we can concatenate both the \emph{to} and \emph{from} dissimilarity matrices to obtain a $2N$-dimensional extended asymmetric dissimilarity space~\cite{calana2012using}.

Furthermore, if the identity property is no longer a requirement, the first step in computing $d(B_i,B_j)$ does not need to be the minimum. For instance, we could measure the average of \emph{all} instance distances between two bags, which is rather similar to the set kernel in ~\cite{gartner2002multi}: $K(B_i,B_j) = \sum_{\mathbf{x}_{ik} \in B_i, \mathbf{x}_{jl} \in B_j} k(\mathbf{x}_{ik}, \mathbf{x}_{jl})$ where $k$ is a polynomial, radial basis or other type of kernel on feature vectors. The dissimilarity version is:

\begin{equation}
d_{meanmean}(B_i, B_j)  =  \frac{1}{n_i n_j}\sum_{k=1}^{n_i}\sum_{l=1}^{n_j} d(\mathbf{x}_{ik}, \mathbf{x}_{jl}). 
\label{eq:meanmean}
\end{equation}

\subsection{Bags as Instance Distributions}

%parametrized
Alternatively, we can view each bag as a probability distribution in instance space, and define a bag dissimilarity as a distribution distance. Because the true distributions are not available, we have to approximate the instance distributions, and provide distances between the approximated distributions. 

\begin{figure*}[ht]
\centering
\subfloat[]{
  \includegraphics{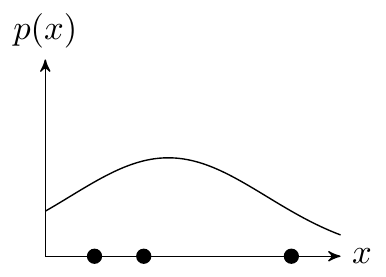}	}
\subfloat[]{
	  \includegraphics{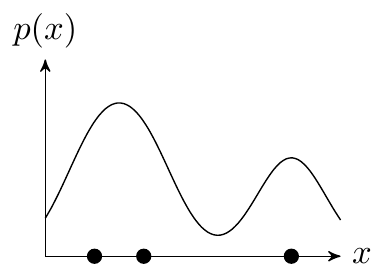}
	}	
\subfloat[]{  \includegraphics{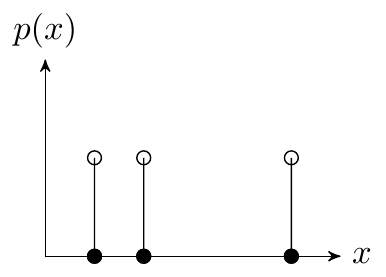}
	}	
\caption{Different ways to represent a 1-D bag with instances at $x=0.5$, $x=1$ and $x=2.5$ as a distribution. From left to right: normal distribution, Parzen density, $\delta$-peaks. The type of approximation influences the choice of distribution distance that can be applied.}
\label{fig:distributions}
\end{figure*}

Fig.~\ref{fig:distributions} shows a number of these approaches to approximate a 1D distribution. These approaches can be seen as a Parzen density with a very large, intermediate, and very small width parameter $\sigma$, resulting in a single Gaussian, an ``intermediate'' multi-modal density, and the empirical density consisting of Dirac deltas. 

The first possibility is to approximate each bag $B_i$ by a normal distribution with parameters $(\mu_i, \Sigma_i)$ and define the bag dissimilarity through the Mahalanobis distance:

\begin{equation}
d_{\text{Maha}}(B_i,B_j) = (\mu_i - \mu_j)^{\intercal} \Bigg( \frac{1}{2}\Sigma_i + \frac{1}{2}\Sigma_j \Bigg) ^{-1} (\mu_i - \mu_j).
\label{eq:maha}
\end{equation} 

Approximating each bag by a normal distribution may, however, be too restrictive. In this case, a multivariate Gaussian (or a Parzen density with a smaller width parameter) can be used instead. The bag dissimilarity measure can then, for instance, be computed as a divergence between the estimated distributions~\cite{budka2011accuracy}. For instance, the Cauchy-Schwarz divergence, rewritten in our notation, can be estimated as:

\begin{align}
d_{\text{CS}}(B_i,B_j) &=  -\log \Bigg( \frac{ K_{\sigma_i+\sigma_j}(B_i,B_j)}{ (K_{2\sigma_i}(B_i,B_i) K_{2\sigma_j}(B_j,B_j))^{\frac{1}{2}} }  \Bigg)\nonumber \\
\text{ where } &  \nonumber \\
K_{\sigma}(B_i,B_j)   &=  \sum_{\substack{\mathbf{x}_k\in B_i\\\mathbf{x}_l \in B_j}} \frac{ \exp\Big(\frac{1}{-2\sigma^2}(\mathbf{x}_k -\mathbf{x}_l )^{\intercal}(\mathbf{x}_k-\mathbf{x}_l)\Big) } {(2\pi\sigma^2)^{\frac{d}{2}}}.
\label{eq:cs}
\end{align}

We follow the advice of ~\cite{budka2011accuracy} to use a single parameter $\sigma$, i.e., $\sigma_i = \sigma_j$ when the distributions (bags) are from the same data source.  There are similarities between (\ref{eq:cs}) and (\ref{eq:meanmean}). The use of the radial basis function in (\ref{eq:cs}) reduces the influence of larger distances on the bag dissimilarity. In some sense, this also happens in (\ref{eq:meanmin}), where the downscaling of the distances is relative to the instance, however, and not global as in~(\ref{eq:cs}).

%Note that rather than allowing a full covariance matrix $\Sigma_i$ as in Eq.~\ref{eq:maha}, here the covariance matrix for bag $B_i$ is assumed to be $\sigma_i^2I$.

Reducing the width parameter of the Parzen window even further, a distribution can be represented as Dirac deltas at each of the instances. One possible distance for this representation is the earth movers distance (EMD) ~\cite{rubner2000earth}. EMD measures the minimum amount of work to transform one probability distribution $B_i$, or pile of earth, into another probability distribution $B_j$, or hole in the ground. We assume that in a bag with $n_i$ instances, each instance has $1/n_i$ of the total probability mass, so the pile in fact consists of $n_i$ smaller piles, and the hole consists of $n_j$ smaller holes. The ground distances between piles and holes are defined by the (Euclidean) instance distances $d(\mathbf{x}_{ik},\mathbf{x}_{jl})$. The distribution distance is then defined as follows:

\begin{equation}
d_{\text{EMD}}(B_i,B_j) = \sum_{\mathbf{x}_{k} \in B_i, \mathbf{x}_{l} \in B_j}f(\mathbf{x}_{k},\mathbf{x}_{l}) d(\mathbf{x}_{k},\mathbf{x}_{l}) 
\end{equation} 
where $f(\mathbf{x}_{k},\mathbf{x}_{l})$ is the flow that minimizes the overall distance, and that is subject to constraints that ensure that the only available amounts of earth are transported into available holes, and that all of the earth is indeed transported: $f(\mathbf{x}_{k}, \mathbf{x}_{l}) \geq 0$, $\sum_{\mathbf{x}_{k} \in B_i} f(\mathbf{x}_{k}, \mathbf{x}_{l}) \leq 1/n_j$, $\sum_{\mathbf{x}_{l} \in B_j} f(\mathbf{x}_{k}, \mathbf{x}_{l}) \leq 1/n_i$ and $\sum_{\mathbf{x}_{k}\in B_i, \mathbf{x}_{l} \in B_j} f(\mathbf{x}_{k},\mathbf{x}_{l}) = 1$.

\subsection{\added{Contrast with related approaches}}

\added{There are several successful MIL methods in the literature that are related to the methods that we advocate.  This section highlights the differences between our approach and those proposed by others.}  

\subsubsection{Distances}
\added{Citation-$k$NN~\cite{wang2000solving} uses the Hausdorff distance to define a distance between the bags. This distance matrix is used together with a nearest neighbor classifier, where to decide the label of a bag $B_i$, both the bags which are nearest neighbors of $B_i$, as well as bags for which $B_i$ is their nearest neighbor, are used to decide the bag label. However, a nearest neighbor classifier does not use all the information that is contained in the dissimilarity matrix~\cite{pkekalska2002dissimilarity}, and in our previous work we have demonstrated that for MIL, such matrices are more informative when used as features in the dissimilarity space~\cite{tax2011bag}.} 

\begin{figure}[ht]
\centering
	\subfloat[]{
  \includegraphics{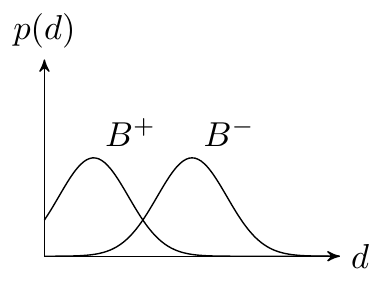}	
  }
	\subfloat[]{
	  \includegraphics{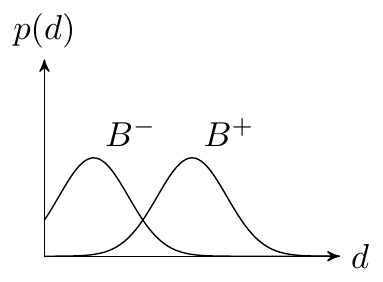}	
	  }	
\caption{Distributions of $d(\cdot, R^{+})$: distances or dissimilarities to a positive prototype bag. Only the situation on the left is suitable for a nearest neighbor approach, while the situation on the right is equally informative for classification in the dissimilarity space.}
\label{fig:sketch}
\end{figure}

Although our approach uses a dissimilarity matrix between bags, it is important to understand that although $d(B_i,B_j)$ can be seen as a bag \emph{distance}, it is not necessarily our goal to arrive at a definition outputs low values for bags of the same class, and high values for bags of different classes. In this work, ``distance'' and ``dissimilarity'' are therefore not equivalent, rather, ``dissimilarity'' is a generalization of ``distance''. Given a certain prototype $R \in \mathcal{R}$, it is sufficient if a dissimilarity produces \emph{discriminative} values for positive and negative bags. For a distance, only the situation on the left of Fig.~\ref{fig:sketch} is acceptable, but for a dissimilarity approach, both situations are equally informative.

\added{For methods such as nearest neighbor, metricity is therefore a desirable property. For the dissimilarity space, this is not the case as we will demonstrate in our Section~\ref{sec:props}}

\subsubsection{Kernels}

In ~\cite{gartner2002multi}, a bag kernel is defined either as a sum of the instance kernels, $K(B_i,B_j) = \sum_{\mathbf{x}_{ik} \in B_i, \mathbf{x}_{jl} \in B_j} k(\mathbf{x}_{ik}, \mathbf{x}_{jl})$ where $k$ is a polynomial, radial basis or other type of kernel on feature vectors, or as a standard (linear or RBF kernel) on instance statistics of each bag. One example is the Minimax representation, where each bag is represented by the minimum and maximum feature values of its instances. 

Similarly to distance-based methods, kernel-based methods, such as such as~\cite{gartner2002multi,zhou2009multi}, also place restrictions on the definition of $K(B_i,B_j)$: a kernel matrix has to be positive semi-definite. In a dissimilarity approach, this is not the case. \added{Any (dis)similarity function can be used, which increases the set of informative similarity functions~\cite{balcan2008theory}}. This is particularly useful if expert advice is to be incorporated in the pattern recognition problem. 

Furthermore, kernel-based methods expect a square matrix, therefore for $N$ training bags, all $N^2$ values need to be available, which is not strictly necessary in a dissimilarity-based approach. In practice we only need dissimilarities to a subset of size $M$ of the training bags. For instance, by choosing $M=log(N)$ and choosing the prototypes beforehand (randomly or by including expert advice), the cost of computing the matrix (and subsequently training the classifier) could greatly be reduced, while still producing good performances~\cite{pkekalska2006prototype}. 

%An alternative to random selection of prototypes is performing a clustering, and then selecting cluster centers as the prototypes, as in BARTMIP~\cite{zhang2009multi2}. However, the clustering already requires a full dissimilarity matrix, so computational effort would only be saved in training the classifier. Furthermore, as explained earlier, bag dissimilarities do not always behave as proper distances. Therefore it is unclear whether such clustering results would be meaningful and that the selected prototypes are better than a random subset of the bags. 

\subsubsection{Instance similarity}
\added{MILES~\cite{chen2006miles} uses a similarity function between bags and all instances in the training set to create a high-dimensional representation for bags. A sparse 1-norm SVM is then used to simultaneously maximize the bag margin, and select the most discriminative similarities, i.e., instances. Conceptually this approach is similar to ours, however there are crucial differences:}

 \begin{itemize}

\item \added{Using bags, rather than instances, as prototypes, significantly reduces the dimensionality. Furthermore, considering instances in a bag jointly helps to capture interactions between instances, which are lost when considering each instance independently.}

\item \added{Using a radial basis function as similarity, assumes that the informative instances are found in clusters, and that only small distances (inside a cluster) can be informative, as sufficiently large distances will be set to zero by the kernel. In cases where the cluster assumption is not correct, using the distances directly still leads to informative directions in the dissimilarity space.}
\end{itemize}

\begin{figure*}[ht]
 \centering
 \subfloat[Concept]{
  \includegraphics[width=0.33\textwidth]{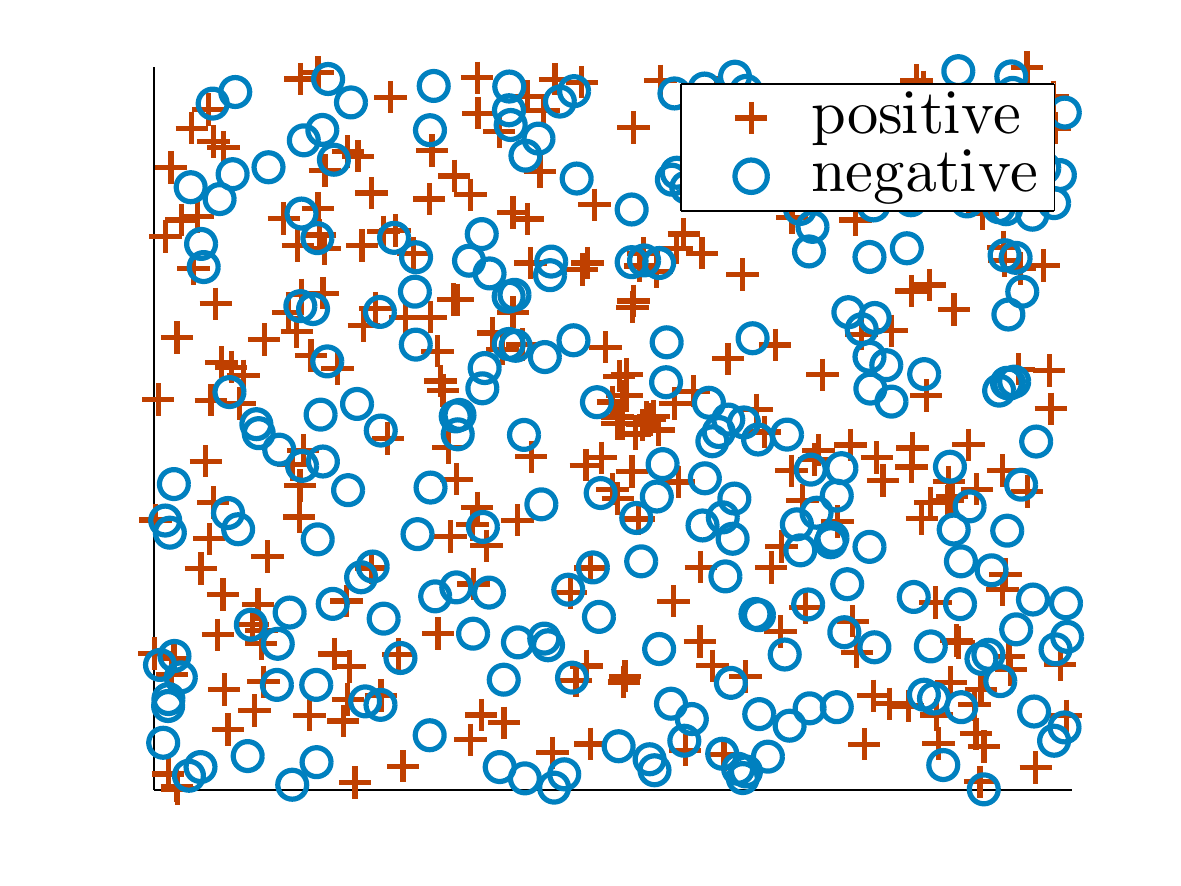}
  }
 \subfloat[Distribution]{
  \includegraphics[width=0.33\textwidth]{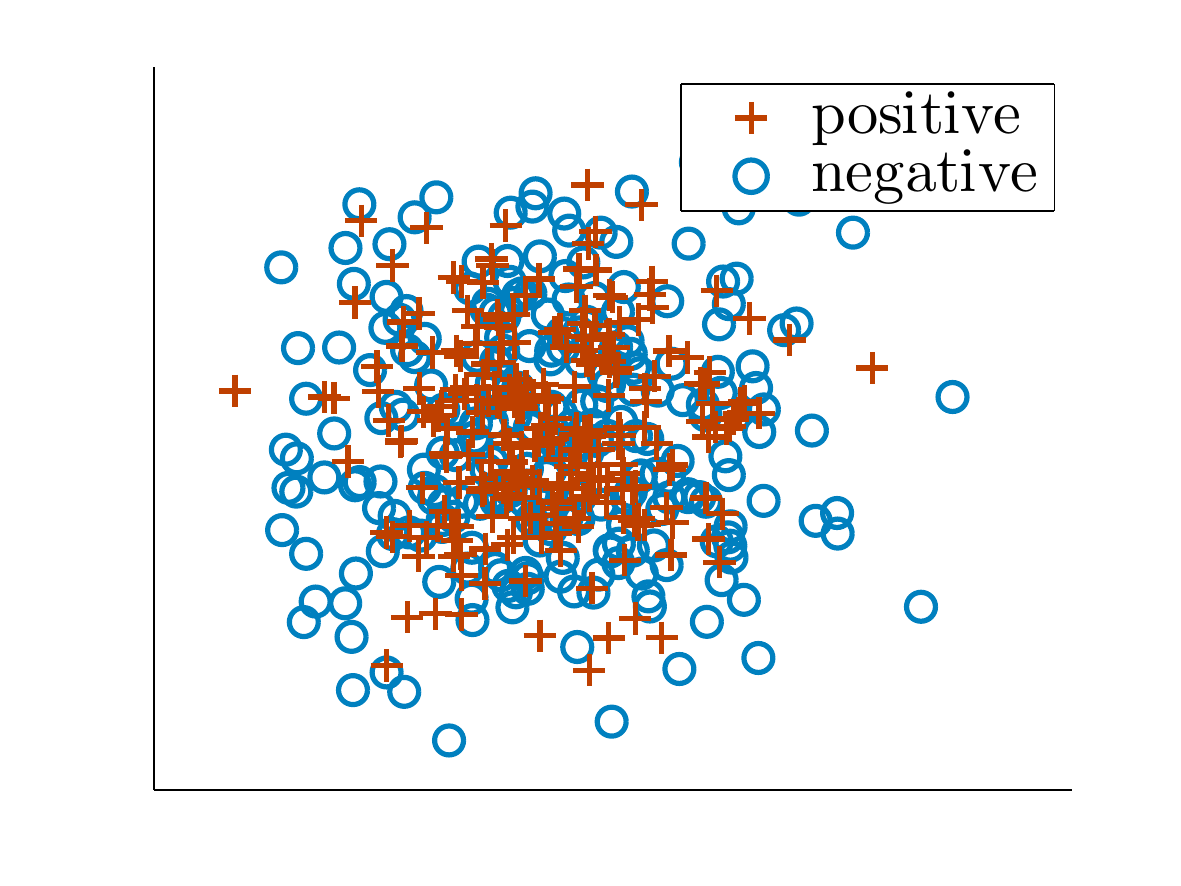}
}
  \subfloat[Multi-concept]{
  \includegraphics[width=0.33\textwidth]{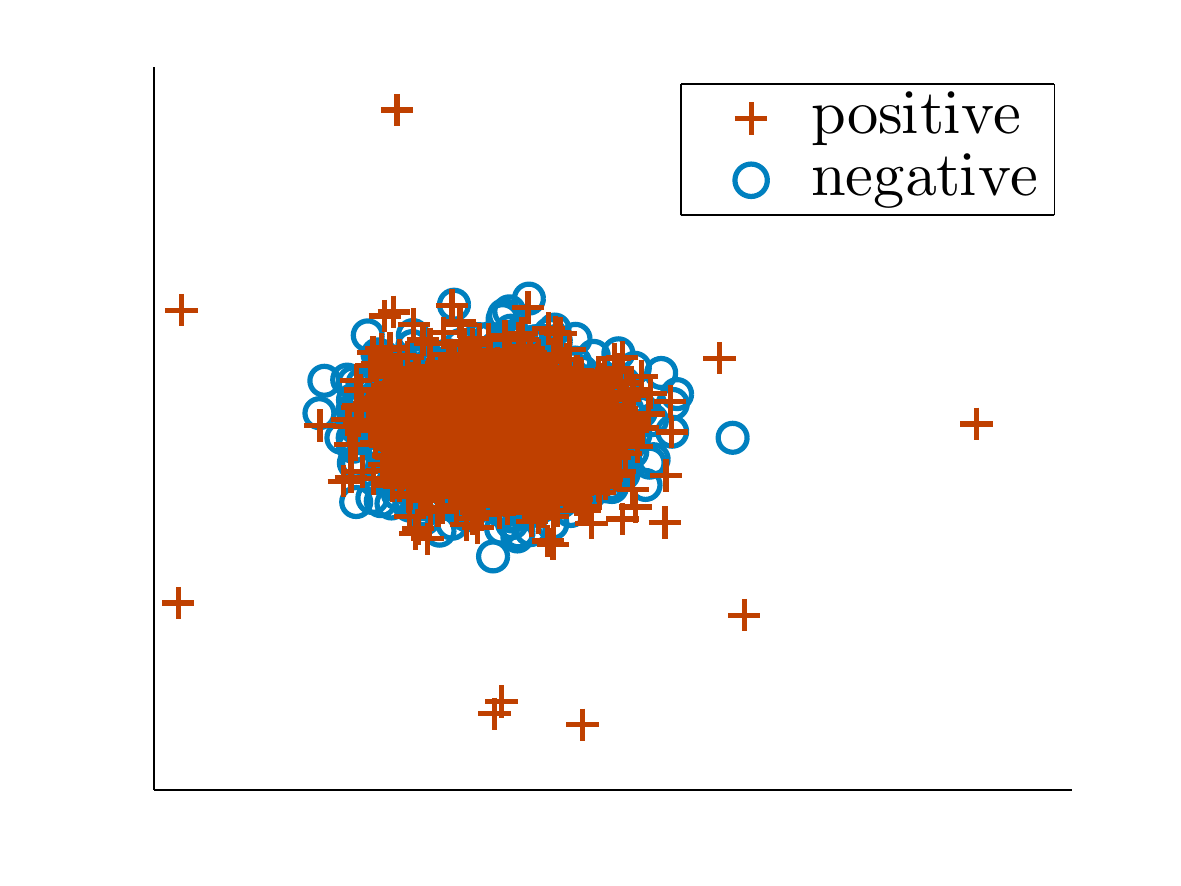}
 } 
 
 \caption[]{Artificial datasets Concept (C), Distribution (D) and Multi-concept (M). The informative instances for these datasets are as follows: C - only the positive instances in the middle, D - all the instances, M - the outlying instances. In the plots, $+$ and $\circ$ are instances of positive and negative bags respectively.}
 \label{fig:examples}
\end{figure*}

\added{To illustrate these points more clearly, we introduce some artificial datasets in Fig.~\ref{fig:examples}. The Concept dataset~\cite{maron1998framework} is a traditional MIL dataset where positive bags have 1 concept instance and $(S-1)$ background instances, and negative bags have $S$ background instances. Due to the dense concept in the middle, (dis)similarities to the concept instances are informative. MILES offers advantages in the Concept dataset, because only a few instances are informative, and each positive bag has such an informative instance.}

\added{The Distribution dataset also has bags with $S$ instances each. Here, the bag as a whole is a more discriminative source of information than a particular instance, because the distributions overlap. Therefore, it is better to consider the instances in each bag jointly, as in our approach. On the other hand, MILES may try to select the negative instances that least overlap with the positive class as informative. However, because the sparsity (number of selected instances) cannot be controlled explicitly, and the high dimensionality, MILES might select too few instances to classify the bags correctly. Furthermore, the non-zero coefficients (the selected instances) may be unstable, leading to high variance in performance if the training set changes.} 

\added{In the Multi-concept dataset, there are also several possible concepts outside the main concentration of instances, but only one of these concepts needs to be satisfied for the bag to be positive. If only a small number of bags is available, it may be the case that each of the concepts is satisfied only by one bag. This bag's concept instance, $\mathbf{x}_c$, is then at a relatively large distance to all other instances. The similarity to $\mathbf{x}_c$ will be set to zero for all bags, turning an informative instance into a non-informative feature. On the other hand, using the distance of $\mathbf{x}_c$ directly creates a feature that reflects the property ``positive bags have an instance that is at a large distance from all bags'', which is very informative for this data.}

\section{Multiple Instance Datasets}\label{sec:behave}

We have gathered a range of datasets, where the tasks vary from image classification to text categorization to predicting molecule activity -- all applications where multiple instance learning has been thought to be beneficial. We believe that, due to the way the datasets are generated (feature extraction, sampling of instances, sampling of bags), not all of these datasets may behave in the same way as MIL has originally been defined. We therefore attempt to cover as many of such situations as possible in order to examine how different MIL methods perform in each case. 
A list of dataset properties is shown in Table~\ref{tab:data}, the datasets themselves (in MIL toolbox~\cite{MIL2011} format) can be found on our website \website.

\begin{table}[h]
\centering

\begin{tabular}{l   p{0.95cm} p{0.95cm} p{0.95cm} p{1cm} p{1cm} p{1cm} p{1cm} }

Dataset & +bags & -bags & dim & inst & avg & min & max \\
\hline
Musk 1  & 47 & 45 & 166 & 476 & 5 & 2 & 40  \\ 
Musk 2 & 39 & 63 & 166 & 6598 & 65  & 1 & 1044 \\
Fox  & 100 & 100 & 230 &1302 & 7  & 2 & 13 \\
Tiger & 100 & 100 & 230 & 1220  & 6 & 1 & 13 \\
Elephant  & 100 & 100 & 230  &1391  & 7 & 2 & 13\\
%Mutagenesis 1 & 125 & 63 & 10486 & 56  \\
African & 100 & 1900 & 9 & 7947 & 8 & 2 & 13\\
Beach& 100 & 1900 & 9  & 7947 & 8 & 2 & 13\\
AjaxOrange & 60 & 1440 & 30 & 47414 & 32 & 31 & 32\\
Web1 & 21 & 92 & 5863 & 2212 & 29 & 4 & 131 \\
Web4 & 87 & 26  & 5863 & 2212 & 29 & 4 & 131 \\
Alt.atheism  & 50 & 50 & 200 & 5443  & 54 & 22 & 76  \\
Comp.graphics & 49 & 51 & 200 & 3094 & 31 & 12 & 58 \\
Brown Creeper & 197 & 351 & 38 & 10232  & 19 & 2 & 43 \\
Winter Wren  & 109 & 439 & 38  & 10232 & 19 & 2 & 43 \\
\end{tabular}
\caption{\added{MIL datasets, number of bags, dimensionality, number of instances and the average, minimum and maximum number of instances per bag. The datasets are available online at \texttt{http://www.miproblems.org}}}
\label{tab:data}
\end{table}

%\subsubsection{Molecule Activity}
The Musk datasets~\cite{dietterich1997solving} are molecule activity prediction problems, where bags are molecules and instances are different conformations of these molecules. The shape of a conformation is responsible for its binding properties, the feature vectors therefore describe the surface properties of the conformations. The standard MIL assumption holds here: as soon as at least one of the conformers has a musky smell, the molecule is classified as having the ``musky smell'' property.

African and Beach are both from the Corel scene classification data~\cite{chen2006miles}. It has been pointed out~\cite{chen2006miles} that, for categories such as ``Beach'', there is not just a single concept: both ``water'' and ``sand'' probably need to be present. This means that the assumption that a single positive instance is sufficient to determine the bag label is not correct. For other categories, it is even difficult to imagine what the concepts might be. For instance, the class ``Historical buildings'' contains images of the ancient Greek and Roman structures, but also of the interiors (floors, staircases) of buildings from a much later period, that do not seem to share any of the same visual queues.

The AjaxOrange dataset originates from the SIVAL~\cite{rahmani2005localized} data. The original contains images of 25 different objects, such as a bottle of dish soap or an apple, and each object is shown in front of different backgrounds, with varying orientation and lighting conditions. Ideally, the whole object should be the concept. However, some objects, such as the bottle, are difficult to segment, causing several concepts (parts of that object) to emerge. Although this may seem similar to the ``multiple concepts'' situation as in scene classification, here orientation and lighting conditions may influence which of the object parts are actually seen in the image. 

In the bird song datasets Brown Creeper and Winter Wren~\cite{briggs2012acoustic}, a bag is an audio fragment consisting of bird songs of different species. Whenever a particular species is heard in the fragment, the bag is positive for that category. It could be expected that birds of the same species have similar songs, therefore there should be different concepts for different bird species. It is also possible that some species are heard together more often. Indeed, there are correlations up to 0.7 between the labels of some species. In this case, instances which are negative for one species, could still be helpful in classifying fragments as containing that species or not.%, which is certainly not the case in the artificial Concept dataset. 

In Newsgroups~\cite{zhou2009multi}, a bag is a collection of newsgroup posts or messages. At the first glance, it seems that this is a typical Concept-type dataset: a positive bag for the category ``politics'' contains 3\% of posts about politics, whereas negative bags contain only posts about other topics. What is different here, is that posts about politics may have nothing in common and thus be very far apart in the feature space. In other words, there are several concepts, but it is sufficient to satisfy only one of these concepts. The Multi-concept dataset may be a reasonable approximation for this dataset. 

%, unlike the concept instances in the artificial Concept dataset. Furthermore, posts from different categories may be very similar: consider a general reply such as ``I do not agree with that opinion''. 
In Web Recommendation~\cite{zhou2005multi}, a bag is a webpage, and instances are other webpages that the first webpage links to. The task is to predict whether to recommend a particular website (bag) to a user based on the linked content, or not. The websites in each of the datasets are the same, but the labels are different for each user that gave his or her preferences. This suggests that here, too, there might be multiple concepts (e.g. content about cooking, and content about sports), but that not all of these need to be satisfied in order for a user to like a webpage. In other words, a user would probably like webpages that link to either cooking, or sports, or both. \added{The dataset has train and test sets defined, for our purposes we concatenated these datasets and use all the data in cross-validation.}

%In TREC~\cite{andrews2002multiple}, a bag is a medical article, and instances are passages from the article. Originally, the articles are annotated with medical terms, each of these terms forms its own category. Each of the seven datasets consists of 100 articles from a certain category, and 100 randomly sampled articles from other categories. A curious fact about TREC is that it was introduced in the same paper~\cite{andrews2002support} as the Fox, Tiger and Elephant datasets. However, since then, the images are almost always used in comparisons by other papers about MIL, while TREC is almost never used. We suspect that TREC might have some properties (such as several concepts) that cause some MIL methods to fail.

\section{Experiments}\label{sec:results}

We start with a comparison of the proposed bag dissimilarities in Section~\ref{sec:comparisonbag}. In Section~\ref{sec:props}, a more in depth analysis of the characteristics of the dissimilarities of real world data is given. A subset of these bag dissimilarities is then compared to other MIL approaches in Section~\ref{sec:comparisonmil}. The metric used for comparisons is area under the receiver-operating characteristic (AUC)~\cite{bradley1997use}. This measure has been shown to be more discriminative than accuracy in classifier comparisons~\cite{huang2005using}, and more suitable for MIL problems~\cite{tax2008learning}.

\added{For intermediate experiments, we use a subset of the datasets from Table~\ref{tab:data} because datasets from the same source (such as one-against-all datasets with a different positive class) are expected to show similar behaviour. For the final comparison, all the datasets are used.}

In this section, the notation $D$ is used to denote the full dissimilarity matrix, as opposed to $d$, which stands for a single dissimilarity value between two bags.
 
\added{Each experiment is performed using cross-validation, where in each fold, the bags are split up into $N_{tr}$ train and $N_{te}$ test bags. The train bags are then used as prototypes to compute the $((N_{tr} + N_{te}) \times N_{tr})$ bag dissimilarity matrix $D$, i.e., no prototype selection is performed. The only dissimilarity measure where a parameter needs to be set is $d_{CS}$. Here we used a default value, square root of the dimensionality) for all the datasets. It is therefore possible that these results could be improved, however at the added cost of cross-validating over different values for $\sigma$.}

\added{In this dissimilarity space, any supervised classifier can be applied; in this paper we use the logistic and support vector (SVM) classifiers. On average, the support vector classifier results were superior to those of the logistic classifier, therefore in some of the further experiments, only SVM results are reported. The classifiers are used with default parameters: regularization parameter $C=1$, and the SVM is used with a linear kernel unless stated otherwise.}

\subsection{Point Set Dissimilarities}~\label{sec:comparisonpointset}

The first comparison is between the bag dissimilarities that are most closely related to the Hausdorff distance: the symmetrized versions of $D_{minmin}$, $D_{meanmin}$ and $D_{maxmin}$. The artificial datasets Concept, Distribution and Multi-concept from Fig.~\ref{fig:examples}, denoted by C, D and M respectively, are very suitable to demonstrate the strengths and weaknesses of these bag dissimilarities. The success of a bag dissimilarity is determined by whether it allows the informative instances (those that cause the differences between positive and negative bags) to sufficiently influence the dissimilarity value. Because the overall mean dissimilarity in (~\ref{eq:meanmean}) is naturally symmetric, we also include it in this comparison.

Table ~\ref{tab:perf1a} shows performances of the bag dissimilarities for the artificial (at two different sample sizes of 25 and 50 bags per class) and seven real-life MIL problems. %For the real-life problems, we mostly use one dataset per group of datasets, in particular Corel African, SIVAL AjaxOrange, alt.atheism newsgroup, Brown Creeper birdsong and Web recommendation 1. The datasets within such a group are often not independent, because they originate from the same multi-class dataset. As some groups are larger due to original number of classes (20 for Newsgroups, 25 for SIVAL), methods performing well on that particular type of data could to be at an unfair advantage.

\begin{table}[ht]
\caption{Point set, symmetrized dissimilarity, logistic and SVM classifiers. AUC and standard error ($\times100$), $5\times10$-fold cross-validation. Bold = best (or not significantly worse) result per dataset.} %The trend is that minmin and instance-based methods do well in ``Concept'' situations, and meanmin and bag-based methods do well in ``Distribution'' situations.}
\begin{center}
\begin{tabular}{l l r r r r}

%& \multicolumn{4}{c}{representation} \\

Classifier & Data &    $D_{minmin}$ &   $D_{meanmin}$ &  $D_{maxmin}$ &   $D_{meanmean}$ \\ 

\hline

 \multirow{13}{5mm}{\begin{sideways}\parbox{10mm}{Logistic}\end{sideways}} &  C25 & {\bf 61.4 (2.3)}& 54.8 (2.2)& 47.5 (2.2)& 53.4 (2.1)\\
& C50 & {\bf 98.6 (0.4)}& 79.6 (1.9)& 50.3 (3.3)& 65.6 (3.0)\\
& D25   & 86.2 (1.8)& {\bf 97.8 (0.5)}& 96.6 (0.6)& 69.4 (2.4)\\
& D50   & 91.7 (1.1)& {\bf 100.0 (0.0)}& 99.6 (0.2)& {\bf 100.0 (0.0)}\\
& M25   & 54.4 (2.2)& 50.8 (1.9)& 60.5 (2.3)& {\bf 71.4 (1.9)}\\
& M50   & 71.5 (2.4)& 78.4 (2.1)& {\bf 84.3 (1.5)}& 69.4 (2.4)\\

 %\hline 

&    Musk1 & 88.2 (1.8)& {\bf 90.2 (1.7)}& {\bf 91.8 (1.6)}& 84.3 (1.8)\\
&    Musk2 & 92.0 (1.2)& {\bf 92.6 (1.3)}& {\bf 93.3 (1.2)}& 82.8 (1.7)\\
&    African & {\bf 96.3 (0.4)}& 94.4 (0.6)& 94.2 (0.6)& 91.1 (0.7)\\
&    Ajax & 68.4 (1.6)& {\bf 98.1 (0.5)}& {\bf 97.2 (0.7)}& 87.8 (1.1)\\
&    Alt.ath & 49.2 (0.8)& {\bf 88.5 (1.7)}& 83.7 (1.7)& 85.2 (1.8)\\
&    BrCr & 89.6 (0.6)& {\bf 93.6 (0.4)}& 91.1 (0.5)& 82.3 (0.7)\\
&    Web & 69.7 (4.0)& {\bf 77.0 (3.2)}& 66.8 (3.7)& 69.9 (3.3)\\
    
    \hline
    
 \multirow{13}{5mm}{\begin{sideways}\parbox{10mm}{LibSVM}\end{sideways}} &  C25 & {\bf 57.7 (2.4)}& 52.2 (2.6)& 45.9 (2.1)& 40.8 (1.5)\\
& C50 & {\bf 98.6 (0.4)}& 83.9 (2.0)& 46.1 (2.4)& 66.4 (3.0)\\
& D25   & 72.0 (2.4)& {\bf 78.9 (2.0)}& {\bf 82.7 (1.6)}& 68.2 (2.5)\\
& D50   & 92.9 (1.1)& {\bf 100.0 (0.0)}& 99.5 (0.2)& {\bf 100.0 (0.0)}\\
& M25   & 47.0 (2.2)& 50.6 (2.3)& 66.2 (2.3)& {\bf 71.6 (2.0)}\\
& M50   & 72.0 (2.4)& {\bf 78.9 (2.0)}& {\bf 82.7 (1.6)}& 68.2 (2.5)\\

     %  \hline
&    Musk1 & 92.0 (1.2)& {\bf 93.4 (1.2)}& {\bf 93.4 (1.3)}& 88.2 (1.5)\\
&    Musk2 & 94.0 (1.3)& {\bf 95.4 (1.4)}& {\bf 95.3 (1.2)}& 90.3 (1.5)\\
&    African & {\bf 96.6 (0.3)}& {\bf 96.7 (0.4)}& 95.5 (0.5)& 90.1 (0.7)\\
&    Ajax & 71.1 (1.4)& {\bf 98.6 (0.4)}& {\bf 97.8 (0.5)}& 84.0 (1.2)\\
&    Alt.ath & 50.0 (0.0)& {\bf 94.9 (1.0)}& 91.4 (1.1)& 94.2 (1.1)\\
&    BrCr & 87.8 (0.6)& {\bf 95.5 (0.3)}& 92.6 (0.5)& 54.4 (2.4)\\
&    Web & 53.2 (4.8)& {\bf 76.0 (2.7)}& 43.3 (3.6)& {\bf 77.6 (3.3)}\\

\end{tabular}
\end{center}

\label{tab:perf1a}
\end{table}

%\begin{table}[ht]
%\caption{Point set, symmetrized dissimilarity, Logistic and LibSVM classifiers. AUC and standard error ($\times100$), $5\times10$-fold cross-validation. Bold = best (or not significantly worse) result per dataset.}% The trend is that minmin and instance-based methods do well in ``Concept'' situations, and meanmin and bag-based methods do well in ``Distribution'' situations.}
%\begin{center}
%\begin{tabular}{l c c c c}
%& \multicolumn{4}{c}{representation} \\

%Data &    $D_{minmin}$ &   $D_{meanmin}$ &  $D_{maxmin}$ &   $D_{meanmean}$ \\ 

%\hline
    
%\end{tabular}
%\end{center}

%\label{tab:perf1b}
%\end{table}

$D_{minmin}$ dissimilarity is best suited for the Concept dataset, as only the minimum instance distances inside the concept are informative. $D_{meanmin}$ is expected to be best for the Distribution dataset, because most/all of the instances need to be considered to determine the bag label. $D_{maxmin}$ dissimilarity is most suitable for the Multi-concept dataset, as the maximum operation selects the most informative, remote instances.
Although adding more training data benefits most dissimilarities, one of the dissimilarities is doing better than the others. $D_{meanmean}$ performs well on the Distribution and Multi-concept problems it captures the same important characteristics as $D_{meanmin}$ and $D_{maxmin}$ in these cases.

\begin{figure}[ht]
 \centering
 \subfloat{
  \includegraphics[width=0.45\textwidth]{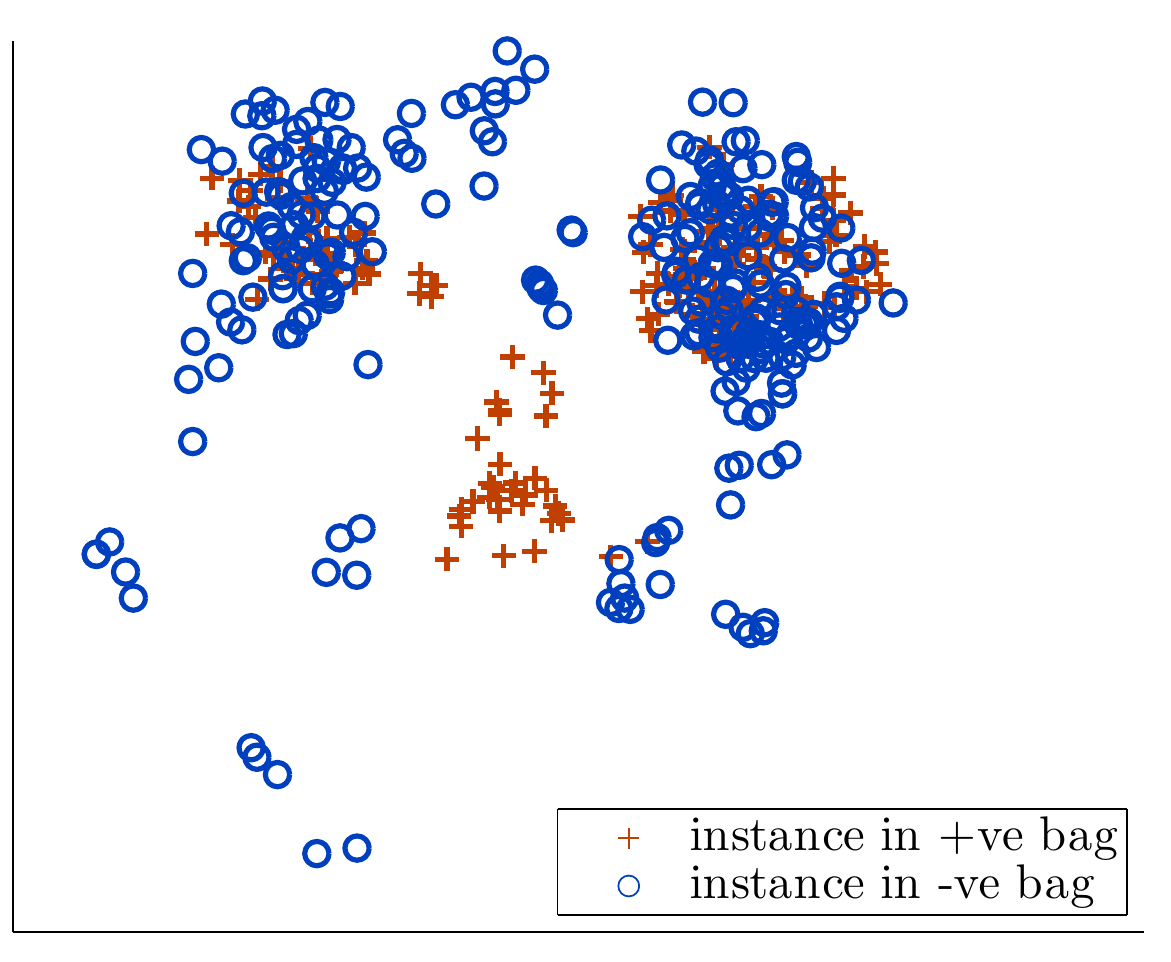}
  }
 \subfloat{
  \includegraphics[width=0.45\textwidth]{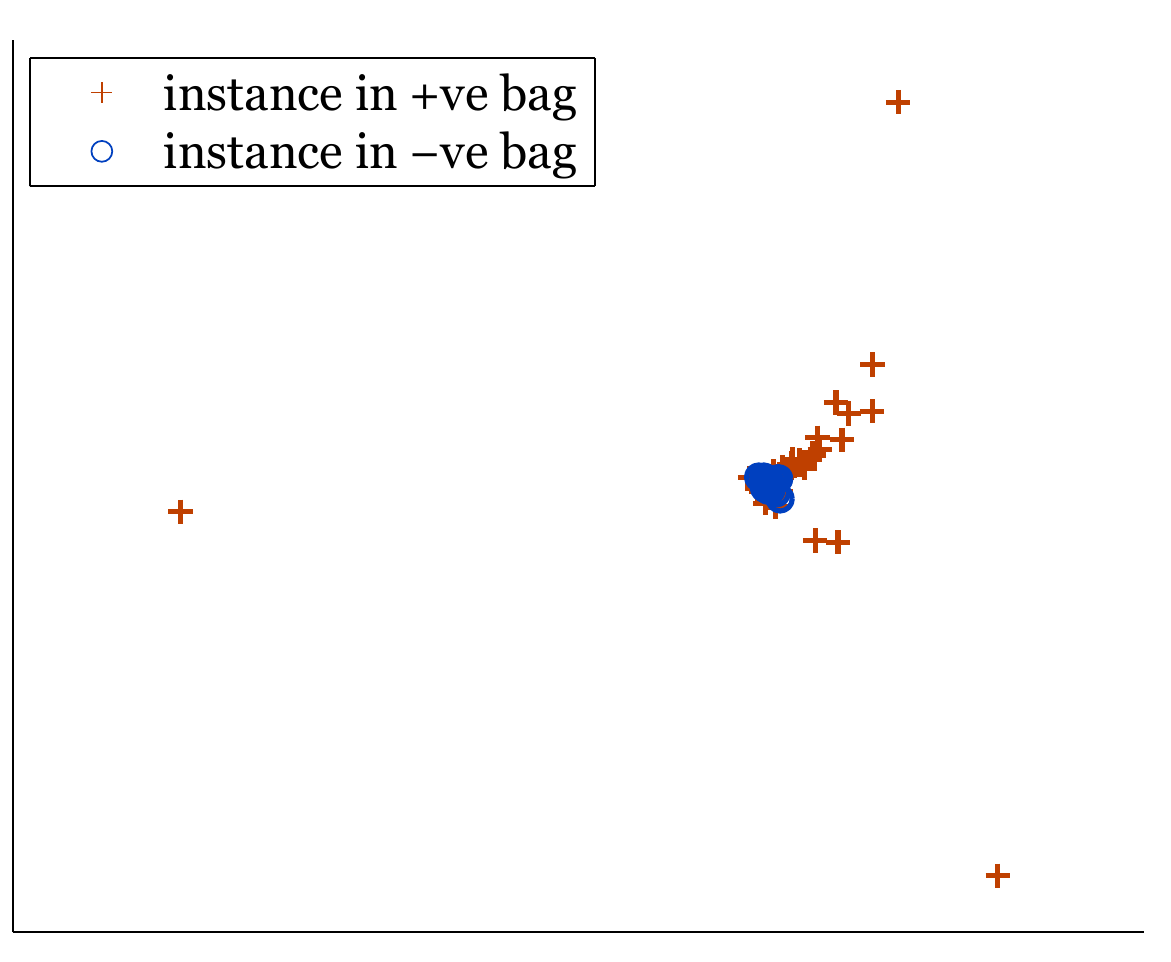}
  }
  \caption[]{\added{Multi-dimensional scaling projection of instances in the Musk 1 (left) and alt.atheism (right) datasets}}
 \label{fig:mds}
\end{figure}

On the real-life datasets, the dissimilarities based on minimum instance distances have comparable performance. This suggests that the distributions of instances from positive and negative bags are different enough that any statistic of the instance distances is able to separate the classes well. \added{Fig.~\ref{fig:mds} shows a 2D projection (obtained by multi-dimensional scaling) of the instances in the Musk 1 dataset. For example, the positive bags that have instances in the center cluster, are slightly closer to each other, than to other bags, under $D_{minmin}$, creating informative dissimilarities. However, other statistics of the instance distances also separate the classes quite well.}

In AjaxOrange and alt.atheism datasets, $D_{minmin}$ performs very badly. In these datasets, positive and negative bags are likely to contain exactly the same instances (background objects in AjaxOrange and very general, uninformative posts in alt.atheism). \added{For example, in the right plot of Fig.~\ref{fig:mds}, we see that the alt.atheism dataset has a cluster of instances in the middle (both from positive and negative bags) and outlying instances from positive bags, similar to the Multi-concept dataset. This is caused by the use of word frequency features: instances containing unusual words are far away from all the others, and instances containing only regular words are very close to most others, i.e., in the middle cluster. Because the instance density in this cluster is very high, $D_{minmin}$ always ``selects'' instances from this cluster.} This results in many bag dissimilarities being equal to zero, creating large class overlap. 

Although the performances on the real datasets are reasonable, $D_{meanmean}$ does not do as well as the other dissimilarities. This suggests that in real problems, the distributions of instances from positive and negative bags might be somewhat different, but that this is not as pronounced as in the artificial Distribution and Multi-concept cases. Therefore, the minimum distance matches for all the instances provide more reliable information about the bag class. \added{An exception is alt.atheism, where the performance of $D_{meanmean}$ is quite high. This can also be explained with the plot in Fig.~\ref{fig:mds}. Here we see outlying positive instances, all belonging to different positive bags, similar to the Multi-concept dataset. These instances are the most informative for the bag class, therefore dissimilarities involving these distances can perform well. This is the case for $D_{meanmin}$, $D_{maxmin}$ and $D_{meanmean}$, which is reflected in the performances}.

Overall, $D_{meanmin}$ gives the best results. Although it is possible that only a few instances per bag are informative (as in the Concept artificial dataset), considering all instances (and their minimum distance neighbors) with equal weight seems to be already be sufficient in practice.

\subsection{Distribution Dissimilarities}

Table~\ref{tab:expdistr} shows the results of the distribution dissimilarities for the logistic and LibSVM classifiers. The result of Musk2 for $D_{\text{EMD}}$ is missing because EMD does not scale well to large bags (some bags in Musk2 have 1000+ instances). The result of Web recommendation for $D_{\text{Maha}}$ is missing because the dataset has almost 6000 dimensions and computing the dissimilarity therefore requires many inversions of $6000\times6000$ covariance matrices. 

Overall, the results of $D_{\text{Maha}}$ are not very good, except for Musk2. A possible reason could be that Musk2 is the only dataset with bags that are large enough for reliable estimation of an inverse covariance matrix. The results of EMD are quite good, however, for these datasets it does not offer clear advantages over the point set dissimilarities. Lastly, the results of $d_{\text{CS}}$ are also reasonable, which is a surprise considering that the results are quite sensitive to the $\sigma$ parameter. 

It is interesting to compare the results of $D_{\text{CS}}$ and $D_{meanmean}$ from the previous section. One of the main differences is that in $D_{\text{CS}}$, more emphasis is put on the smaller distances (due to the RBF kernel), while in $D_{meanmean}$ large distances influence the dissimilarity values significantly. This explains the large gap in the results of alt.atheism, where the outlier instances are very important. On the other hand, $D_{\text{CS}}$ is much better than $D_{meanmean}$ on the Brown Creeper dataset, where a tight concept, and therefore small distances between concept instances, can be expected. 

\begin{table}[ht]
\caption{Distribution dissimilarities. AUC and standard error ($\times100$), $5\times10$-fold cross-validation. Bold = best (or not significantly worse) result per dataset.}
\begin{center}
\begin{tabular}{l l r r r}
Classifier & Data &      $D_{\text{Maha}}$ &  $D_{\text{EMD}}$ &        $D_{\text{CS}}$ \\ 
\hline
 
\multirow{7}{5mm}{\begin{sideways}\parbox{10mm}{Logistic}\end{sideways}} &  
  Musk1 & 70.1 (2.3)& {\bf 88.7 (1.9)}& 84.6 (2.0)\\
  &  Musk2 & {\bf 91.5 (1.3)}& - & 88.8 (1.5)\\
  &  African & 59.9 (1.5)& {\bf 93.3 (0.6)}& 85.9 (1.0)\\
  &  Ajax & 86.9 (1.8)& {\bf 97.9 (0.4)}& 95.5 (0.7)\\
  &  Alt.ath & 49.9 (2.5)& {\bf 84.0 (1.9)}& 59.2 (2.9)\\
  &  BrCr & 63.3 (1.2)& {\bf 94.5 (0.4)}& 89.4 (0.8)\\
  &  Web & - & 69.4 (4.1)& {\bf 75.7 (3.4)}\\
\hline

\multirow{7}{5mm}{\begin{sideways}\parbox{10mm}{LibSVM}\end{sideways}} &  
   Musk1 & 76.5 (2.9)& {\bf 89.8 (1.6)}& {\bf 88.2 (1.7)}\\
  &  Musk2 & {\bf 96.0 (0.9)}& - & 87.4 (1.8)\\
  &  African & 64.8 (1.5)& {\bf 94.7 (0.4)}& {\bf 94.3 (0.5)}\\
  &  Ajax & 87.3 (1.7)& {\bf 98.9 (0.3)}& 98.1 (0.3)\\
  &  Alt.ath & 47.0 (2.3)& {\bf 87.4 (1.7)}& 41.9 (2.5)\\
  &  BrCr & 59.7 (1.1)& {\bf 95.5 (0.3)}& 93.9 (0.4)\\
  &  Web & - & {\bf 77.7 (2.7)}& 69.5 (3.8)\\
    \end{tabular}
\end{center}
\label{tab:expdistr}
\end{table}

\subsection{Properties of dissimilarity matrices}\label{sec:props}

\added{Table~\ref{tab:props} shows several properties of dissimilarity matrices evaluated in our experiments. These properties demonstrate what we have emphasized in Section~\ref{sec:representation}: it is not our goal to arrive at a definition of $d(B_i,B_j)$ that is consistent with one intuition about distance measures, and that dissimilarities that do not behave as distances, may still be informative for classification. The properties we examine~\cite{pekalska2002generalized,distools} are defined as follows:}

\begin{itemize}

%\item PCA 99\%, is the ratio of dimensions needed to explain 99\% of the variance, and the original dimensionality of the matrix. The columns are scaled to variance prior to computation. 

\item \added{NEF stands for negative eigenfraction, and express how well the dissimilarity matrix can be embedded in a Euclidean space. For a Euclidean dissimilarity matrix, the eigenvalues $\lambda_i$ of the corresponding Gram matrix are non-negative~\cite{pekalska2002generalized}. Negative $\lambda_i$ therefore represent non-Euclidean behavior. NEF is defined as $\frac{\sum_{\lambda_i < 0}|\lambda_{i}| }{\sum_{\lambda_j}\lambda_{j}| }$. For a Euclidean distance measure, the value is 0.}

\item \added{NER stands for negative eigenratio, and is defined as $\frac{|\lambda_{min}|}{\lambda_{max}}$. For a Euclidean distance measure, the value is 0.}

\item \added{NMF stands for non-metricity fraction. It is the percentage of triplets $\{D_{i,j}, D_{i,k}, D_{j,k}\}$ that disobey the triangle inequality. For a metric distance measure, the value is 0.}

%\item NN error is the leave-one-out error of a 1-nearest neighbor classifier. 

\end{itemize}

 \begin{table}[ht]
\caption{\added{Properties of dissimilarity matrices: negative eigenfraction (in \%), negative eigenratio and non-metricity fraction (in \%).}}
\begin{center}
\begin{tabular}{l l r r r r r r r}
Measure & Data & $D_{minmin}$ & $D_{meanmin}$ & $D_{maxmin}$ & $D_{mean}$ & $D_{Maha}$ &  $D_{\text{EMD}}$ &  $D_{CS}$ \\ 
 \hline

\hline
\multirow{6}{5mm}{\begin{sideways}\parbox{10mm}{NEF}\end{sideways}} &  
    Musk1 & 31.2 & 25.5 & 22.6 & 21.0 & 48.9 & 27.4 & 27.1 \\
 &   Musk2 & 31.1 & 25.3 & 21.7 & 20.5 & 28.4 & -  & 96.3 \\
 &   Afr & 47.2 & 37.8 & 33.1 & 29.7 & 50.0 & 49.5 & 35.4 \\
 &   Ajax & 47.9 & 32.6 & 41.3 & 40.0 & 49.6 & 29.9 & 32.5 \\
 &   Alt & 49.5 & 21.0 & 22.0 & 93.1 & 36.9 & 37.4 & 9.2 \\
 &   BrCr & 45.5 & 30.7 & 33.6 & 51.1 & 49.9 & 37.3 & 27.3 \\
 &   Web & 18.6 & 4.8 & 3.7 & 25.9 & - & 10.3 & 100.0 \\

\hline 
 
\multirow{6}{5mm}{\begin{sideways}\parbox{10mm}{NER}\end{sideways}} &    
    Musk1 & 0.4 & 0.2 & 0.2 & 0.2 & 1.0 & 0.2 & 0.2 \\
  &  Musk2 & 0.4 & 0.2 & 0.2 & 0.1 & 0.3 & - & 28.1 \\
  &  Afr & 0.7 & 0.4 & 0.3 & 0.3 & 1.0 & 1.0 & 0.3 \\
  &  Ajax & 0.5 & 0.2 & 0.3 & 1.1 & 1.0 & 0.2 & 0.3 \\
  &  Alt & 1.0 & 0.5 & 0.6 & 24.1 & 0.8 & 0.7 & 0.8 \\
  &  BrCr & 0.5 & 0.4 & 0.3 & 0.8 & 1.0 & 0.6 & 0.2 \\
  &  Web & 1.0 & 1.1 & 0.9 & 1.1 & - & 0.5 & 0.0 \\

 \hline   
 \multirow{6}{5mm}{\begin{sideways}\parbox{10mm}{NMF}\end{sideways}} &  
    Musk1 & 5.5 & 2.5 & 1.6 & 0.7 & 27.1 & 3.8 & 5.5 \\
  &  Musk2 & 7.2 & 3.2 & 1.9 & 0.8 & 7.9 & - & 93.3 \\
  &  Afr & 20.1 & 8.1 & 8.8 & 2.3 & 27.8 & 11.4 & 12.4 \\
  &  Ajax & 17.5 & 0.1 & 0.5 & 0.0 & 18.9 & 0.3 & 4.2 \\
  &  Alt & 0.0 & 0.3 & 0.9 & 0.0 & 1.7 & 9.7 & 0.0 \\
  &  BrCr & 7.8 & 0.8 & 2.0 & 4.2 & 26.7 & 3.5 & 13.1 \\
  & Web & 2.4 & 0.0 & 0.0 & 0.0 & - & 0.0 & 97.3 \\ 
  
 \hline

\end{tabular}
\end{center}
\label{tab:props}
\end{table}

\added{One interesting result is for $D_{CS}$ with highly non-Euclidean / non-metric behaviour for Musk2 and Web data. We suspect that the cause of this behaviour is the highly varying bag size in these datasets. The computation of $D_{CS}$ in (6) uses a sum (not an average) of similarity values. Therefore, it is possible that bags $B_i$, $B_j$ and $B_k$, with instances in the exact same locations in the feature space, but where $B_k$ is much better sampled than $B_j$, and $B_j$ is much better sampled than $B_i$, are at different distances to each other. This difference would not picked up by any of the point set dissimilarities, because averaging is employed.  It depends on the application whether this is a favorable property or not. For example, if the bag size is correlated with the bag label, a sum of instance distances could be more informative than an average. This is not the case for Musk 2 or Web, but it could be relevant for other data, for example in Brown Creeper, where the correlation between bag size and bag label is higher than 0.5.} 

\added{Another unusual result is the NEF and NER obtained on the alt.atheism dataset for $D_{mean}$. The positive bags in this dataset contain very outlying instances, as explained in Section~\ref{sec:comparisonpointset}. The dissimilarities of bags with such outlier instances will be very large, which which causes significant problems to embed such objects in a Euclidean space. This causes large negative eigenvalues, and therefore large values for NEF and NER. Of course, this behaviour is also non-metric, however, NMF only measures the amount of dissatisfied triangle inequalities, disregarding the identity requirement of a metric.} 

\added{Overall, these results show that dissimilarity matrices do not need to be Euclidean or metric to still be informative and useful for dissimilarity-based classifiers, which is consistent with the conclusions in~\cite{pkekalska2006non}. Of course, these measures do not take into account the actual labeling, so such properties alone cannot predict the performance of classifiers in the dissimilarity space. However, examining the dissimilarity matrices can aid in understanding more about the data and the different choices of dissimilarity measures.}

\subsection{Comparison to other MIL approaches}\label{sec:comparisonmil}

We have selected several methods, that are often being used in comparisons in recent papers, as algorithms for our own comparison. For a more detailed description of these methods, see Section 2.  We consider $d_{meanmin}$ as the dissimilarity function used in MInD, and LibSVM~\cite{chang2011libsvm} as the classifier used together with this representation. All classifiers are implemented in PRTools~\cite{prtools} and the MIL toolbox~\cite{MIL2011}. Default parameters are used for all cases, unless stated otherwise. The classifiers used are as follows: %these are either parameters recommended by the authors in the original publications or values that lead to reasonable performances on the Musk1 dataset. 

\begin{itemize}

%\item Simple MIL + logistic classifier
\item EM-DD~\cite{zhang2001dd}, with 10 objects used at initialization.
\item mi-SVM~\cite{andrews2002support}, with a linear kernel.
\item MILBoost~\cite{viola2006multiple} with 100 rounds.
\item MILES~\cite{chen2006miles}, with a radial basis kernel.
\item Minimax MI-Kernel~\cite{gartner2002multi} + SVM with a linear kernel. 
\item MInD + SVM with a linear kernel.
\end{itemize}

%[TODO: Maybe show number of user parameters / learned parameters, and what kind of output (instance label, bag label) is possible? Computational complexity?] 

\subsubsection{Learning curves}

The presented classifiers vary significantly in the model assumptions and their complexity. Therefore, we expect that these methods are affected differently by the amount of training data provided, both in terms of classifier performance, as well as computational issues. We provide several learning curves (in the number of training bags) and show the performance and training/testing times of the classifiers. 

The learning curves are generated as follows. For each of the 20 iterations, the dataset is split into 80\% training and 20\% test bags. Then, the training set is subsampled to contain 5, 10, 20 and 40 ~\added{(or the maximum possible number of)} bags per class. This ensures that the test set remains the same for increasing amounts of training data. When the training set is not large enough, the maximum possible number of bags is sampled instead. 

\begin{figure}[ht]
 \centering
 \subfloat{
  \includegraphics[width=0.30\textwidth]{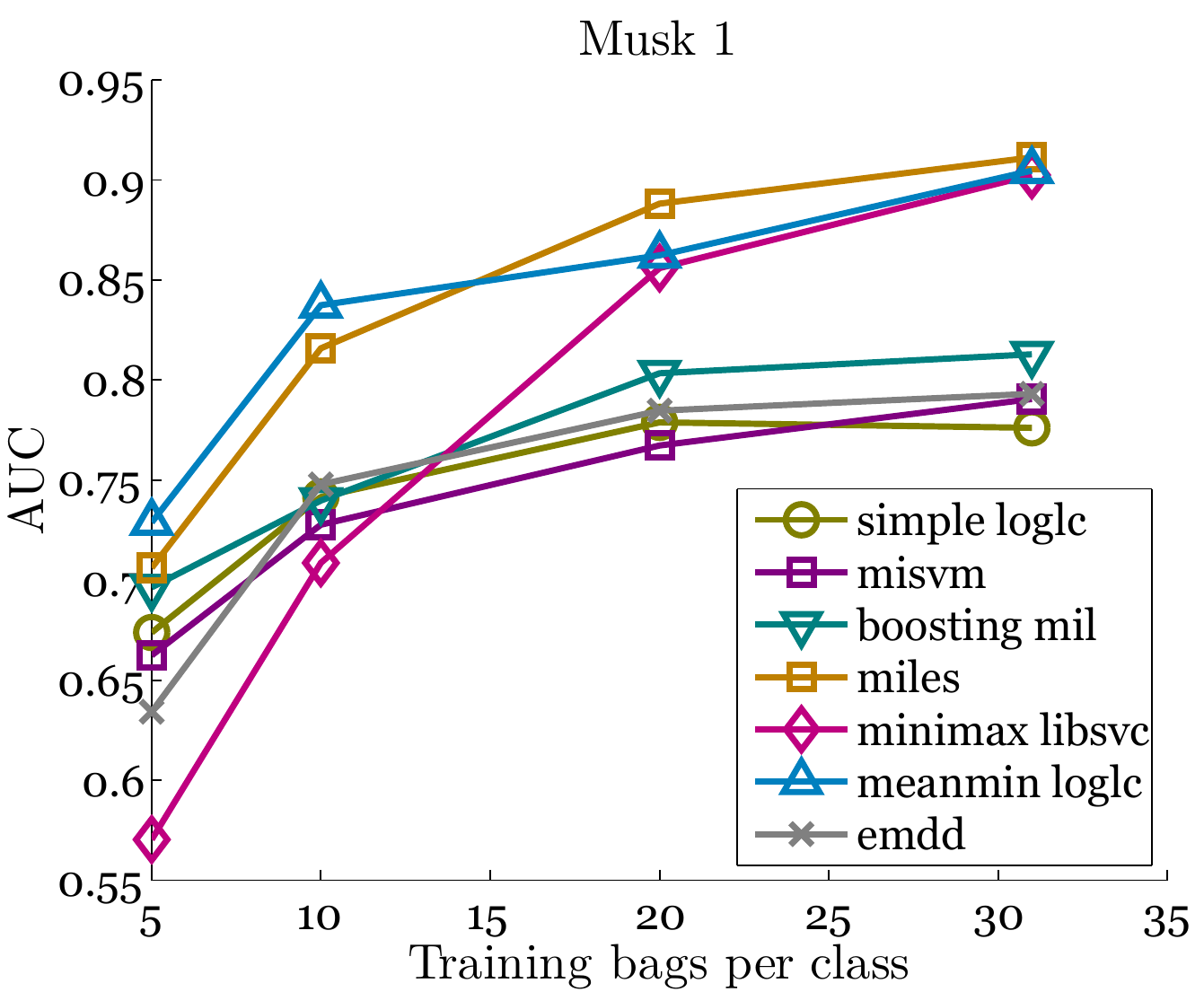}
  }
  \subfloat{
  \includegraphics[width=0.30\textwidth]{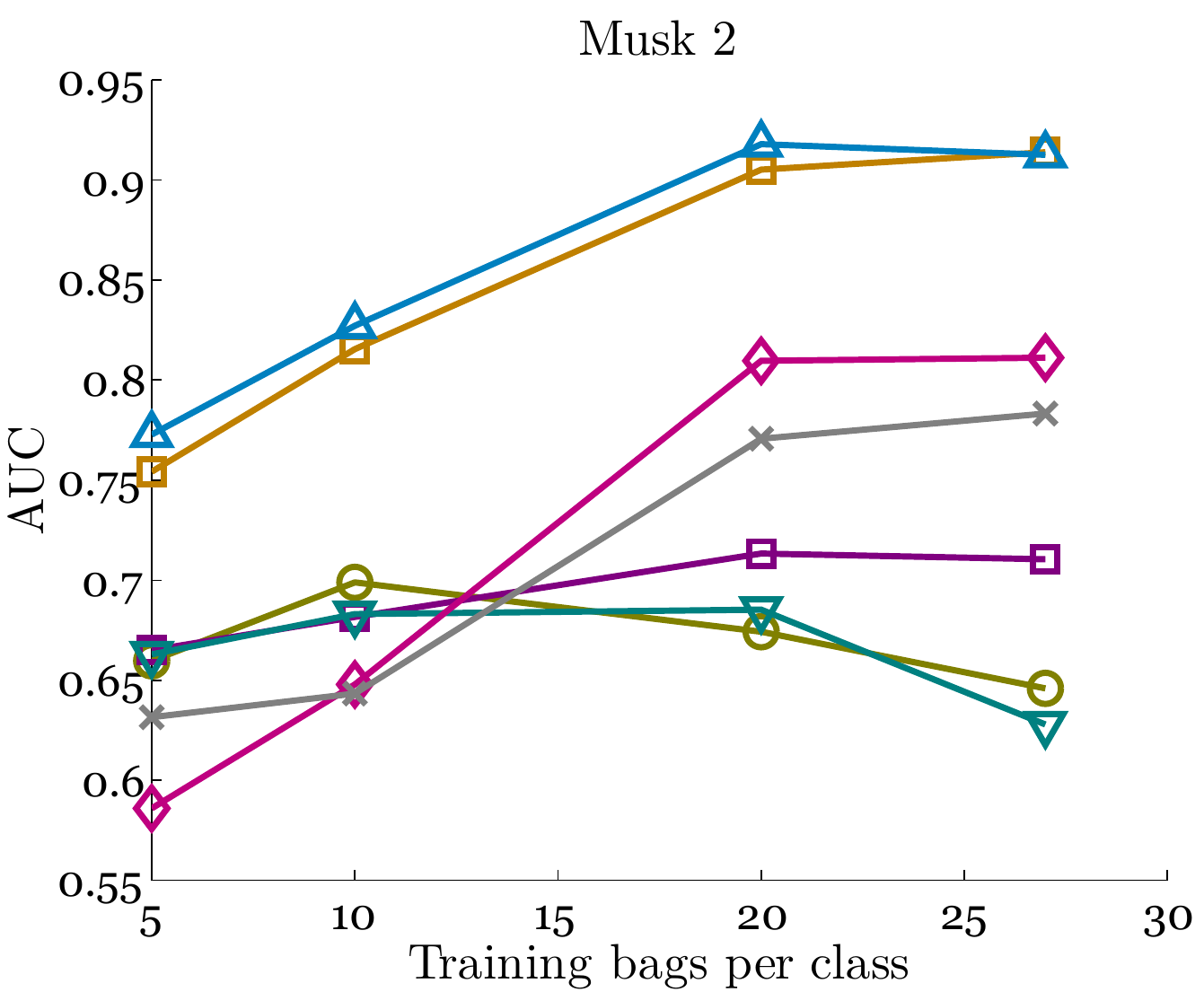}
  }
  \subfloat{
  \includegraphics[width=0.30\textwidth]{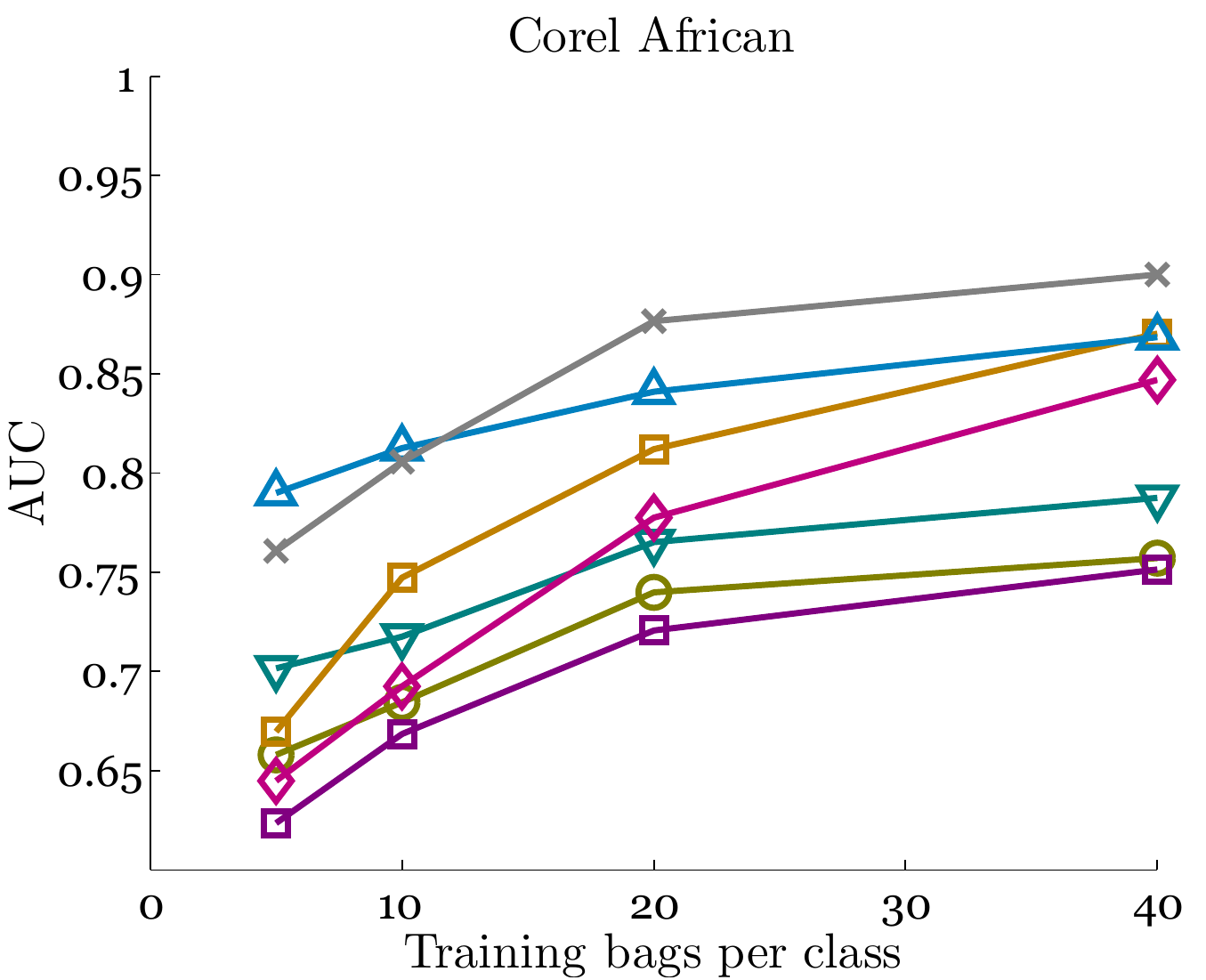}
  }

  \subfloat{
  \includegraphics[width=0.30\textwidth]{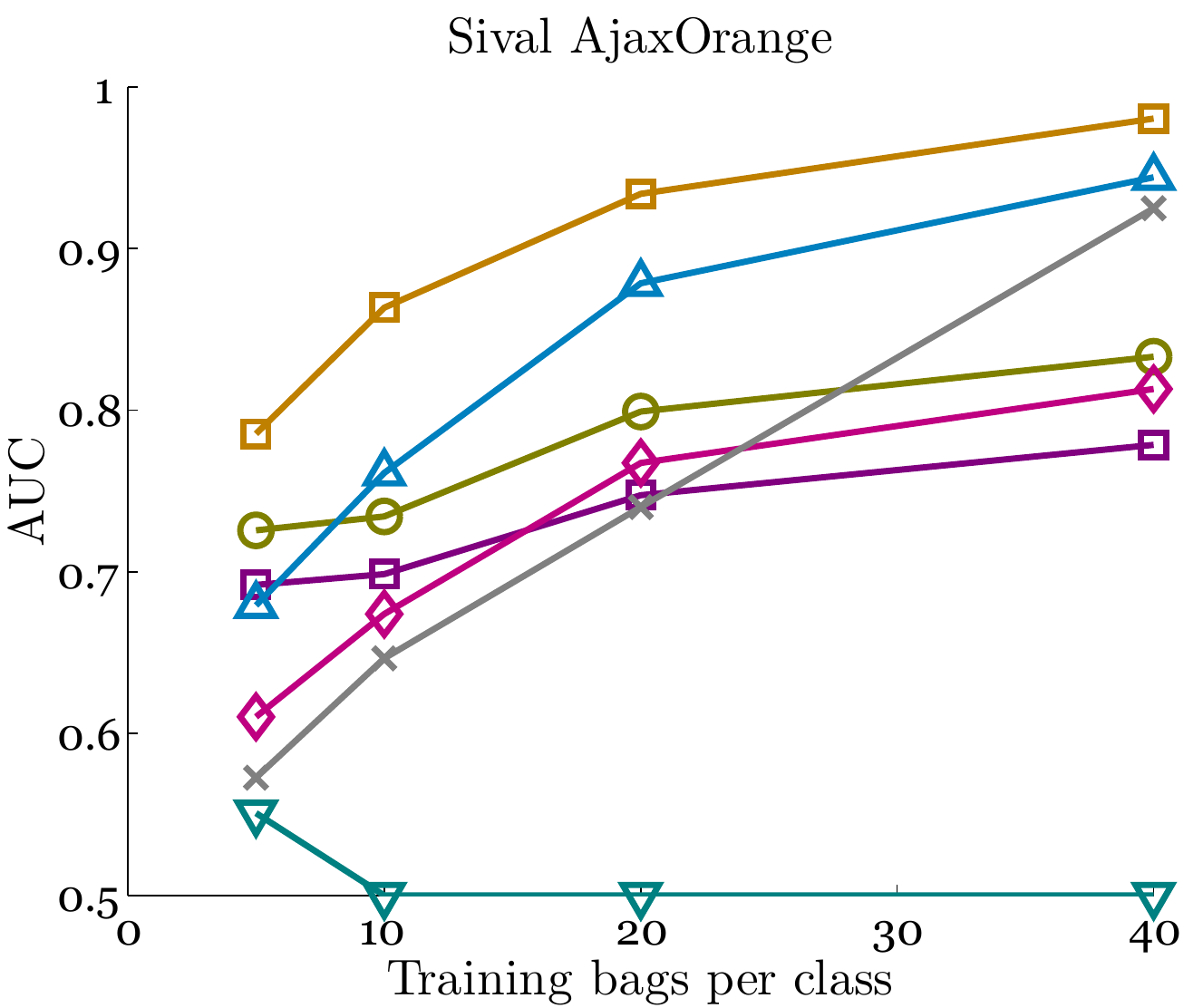}
  }
  \subfloat{
  \includegraphics[width=0.30\textwidth]{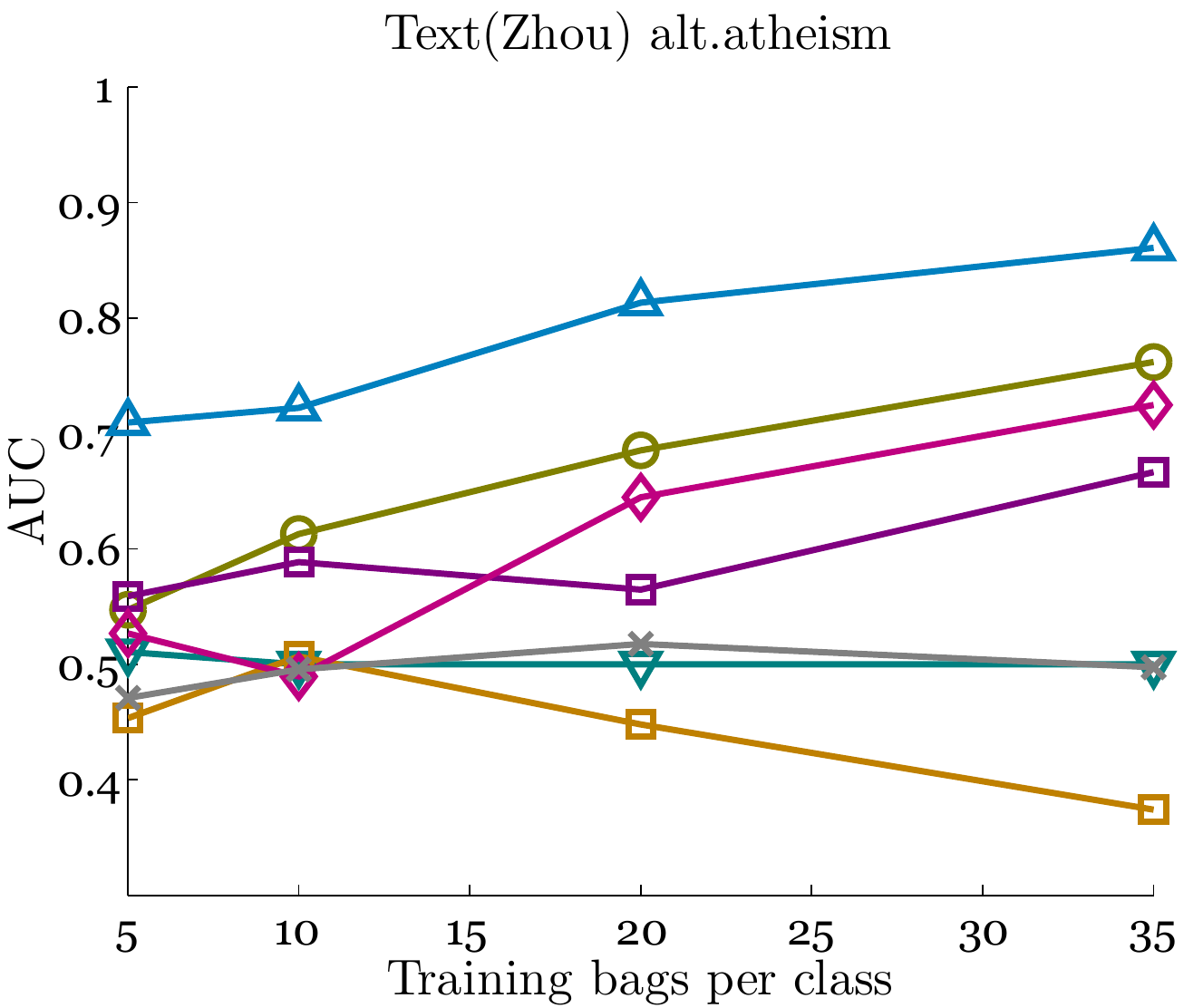}
  }
  \subfloat{
  \includegraphics[width=0.30\textwidth]{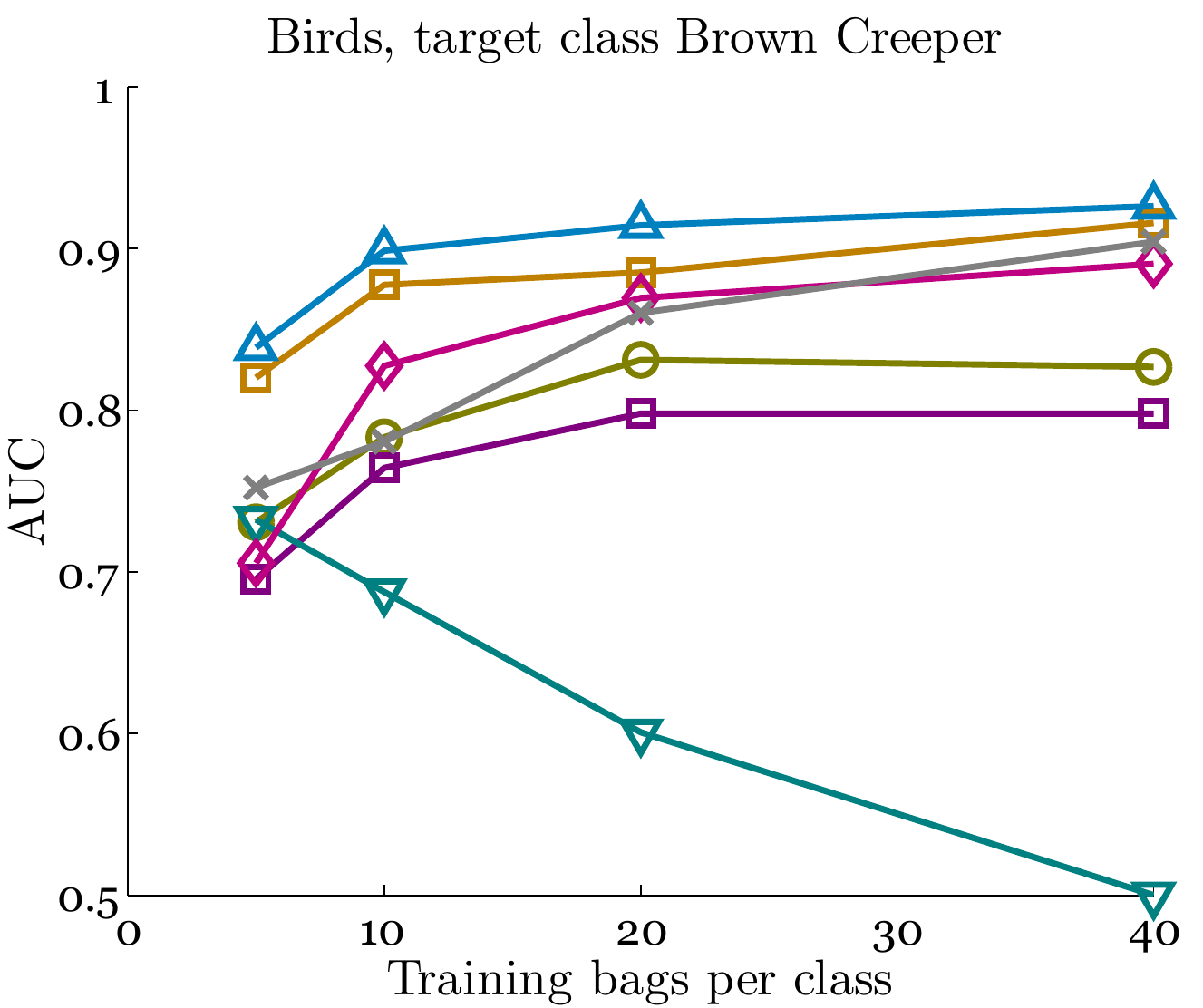}
  }

  \caption[]{Learning curves (AUC performance) for Musk1, Musk2, African, AjaxOrange, alt.atheism and Brown Creeper datasets. The standard deviations are generally around 0.1, but are lower for MILES, Minimax and MInD at larger training sizes. Figure best viewed in color.}
 \label{fig:learncurves1}
\end{figure}

The learning curves for six MIL datasets are shown in Figure~\ref{fig:learncurves1}. In terms of AUC, it is clear that MInD and MILES perform well overall: their performance is always one of the best. There are two exceptions where the results of MInD and MILES are quite different. In the alt.atheism dataset,  MInD outperforms MILES significantly. We suspect that this is because this dataset is similar to the artificial Distribution data - all instances need to be considered to distinguish a positive bag from a negative bag. MInD does this naturally, while MILES tries to enforce sparsity by selecting a few important instances. On the other hand, MILES is superior on the AjaxOrange dataset. This is because MInD takes uninformative background distances into account, while MILES is able to select the instances belonging to the AjaxOrange bottle.  

There is another interesting difference between the MInD and MILES approaches. In all cases except AjaxOrange, where MILES is always superior, the advantage of MInD is more apparent at lower sample sizes. MILES has a higher dimensionality than MInD, which is a disadvantage when only a few bags are available in the training set. \added{This is why the performance of MILES decreases for alt.atheism. As the training set becomes larger, so does the dimensionality, making the problem of selecting informative instances more and more difficult.} 

The Minimax and SimpleMIL approaches also have reasonably good (but typically worse than MILES and MInD) and consistent performances. Both approaches do not require any extra parameters, except the classifier used on top of the representation. EM-DD is able to achieve good results, but only at larger training sizes, as estimating a high-dimensional density requires a lot of data. For MILBoost and mi-SVM, it is not directly clear when a good performance can be achieved. Interestingly, both classifiers try to recover the true instance labels, assuming that at least one instance per bag is positive. As explained in Section~\ref{sec:behave}, this may not be the case for several of the MIL problems examined here. We suspect that this is the reason why methods that learn on bag level and thus make less assumptions, are more successful in our experiments.

%[TODO] Something about decreasing learning curves for MILBoost (AjaxOrange, BrownCreeper)

In terms of time (Figure~\ref{fig:learncurves2}), EM-DD is by far the slowest, followed by MILBoost and mi-SVM due to the optimization of the instance labels. They are followed by MInD and MILES, where creating the (dis)similarity representation is quadratic in the number of instances in the training data. SimpleMIL and Minimax are the fastest methods, because creating the representation is linear in the number of instances.    

When both the performance and training time are taken into account, Minimax, MInD and MILES are the best choices. Although MInD and MILES have higher performances, they are significantly slower. Selecting bags or instances prior to creating the dissimilarity representation could, however, decrease their total training time. Prototype selection after the dissimilarity matrix is already available would increase the training time, but would decrease testing time as less dissimilarities would need to be computed in the test phase. 

Another useful observation is that MILES and bag dissimilarities such as MInD are related because both representations are defined through minimum instance distances. Therefore, the quadratic cost of creating the representation can be shared among these methods, which allows a user to try several competing methods without adding significant computational effort. The same holds for trying several classifiers at a fraction of the cost. In the time curves, the time for computing the representation is included in the time that is displayed, so in practice, these methods would be faster.

%[TODO: Meanmin in these experiments computed twice as many dissimilarities as necessary... Training time can be lowered.]

\begin{figure}[ht]
 \centering
 
   \subfloat{
  \includegraphics[width=0.30\textwidth]{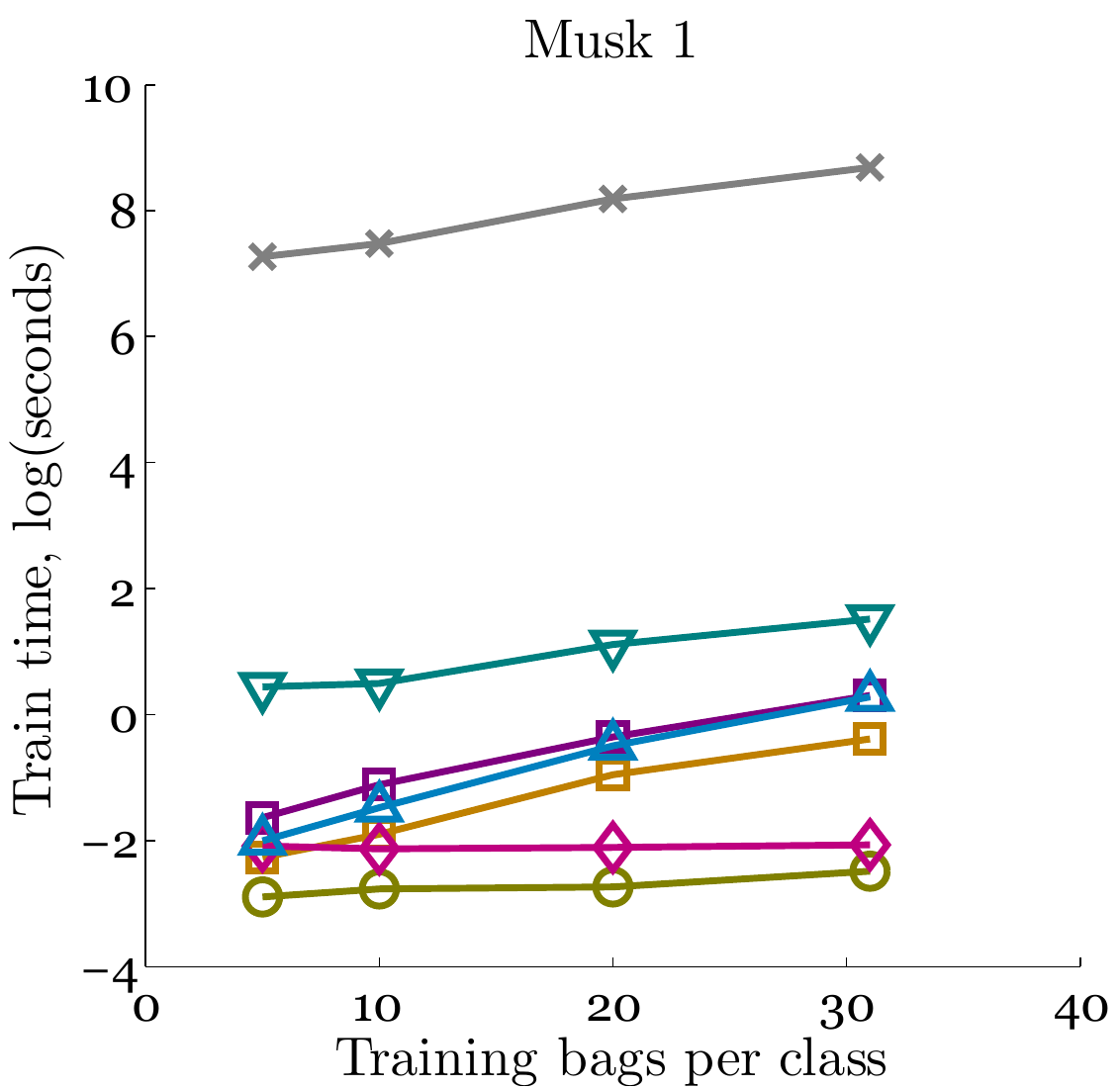}
  }
  \subfloat{
  \includegraphics[width=0.30\textwidth]{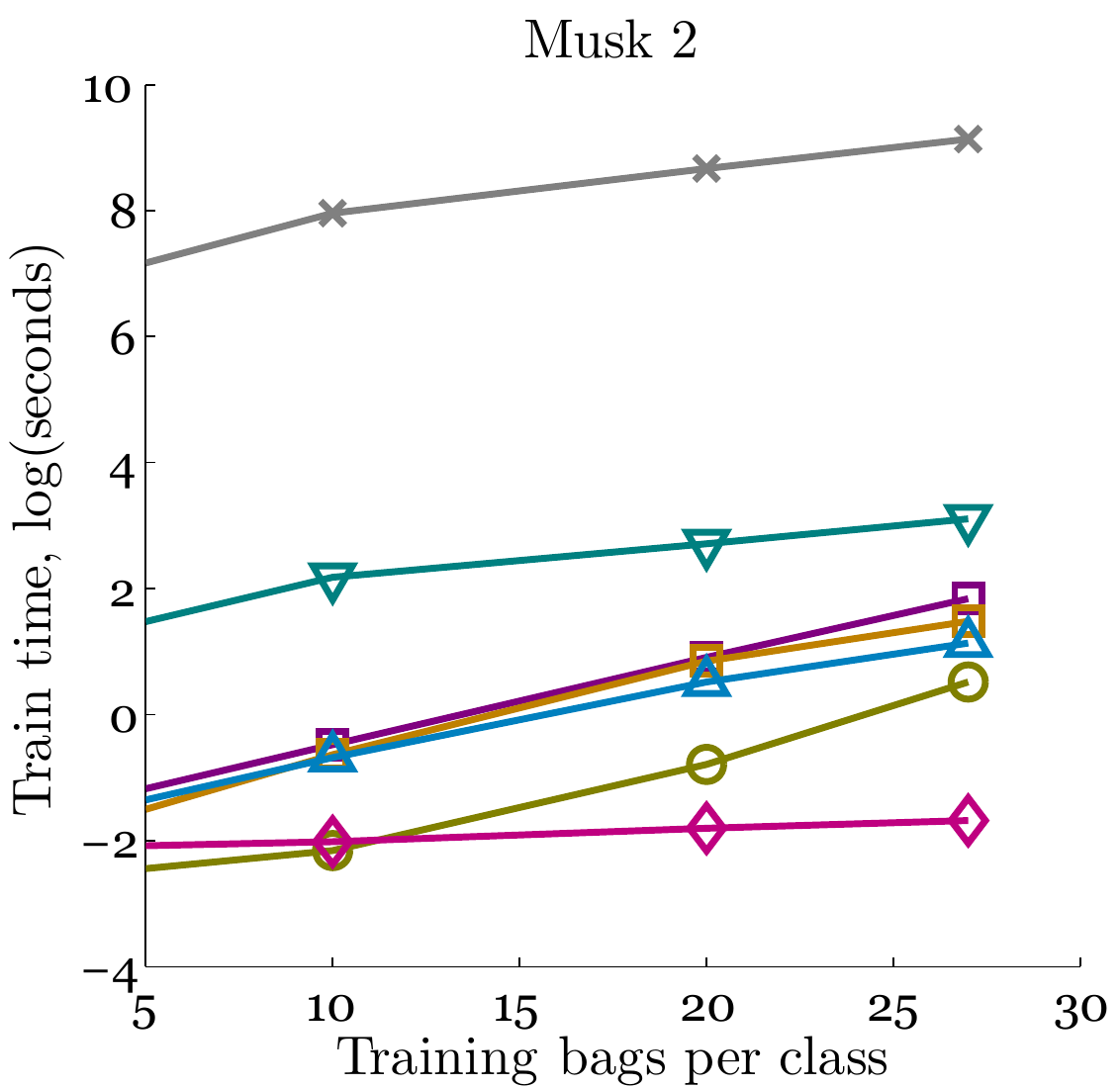}
  }
  \subfloat{
  \includegraphics[width=0.30\textwidth]{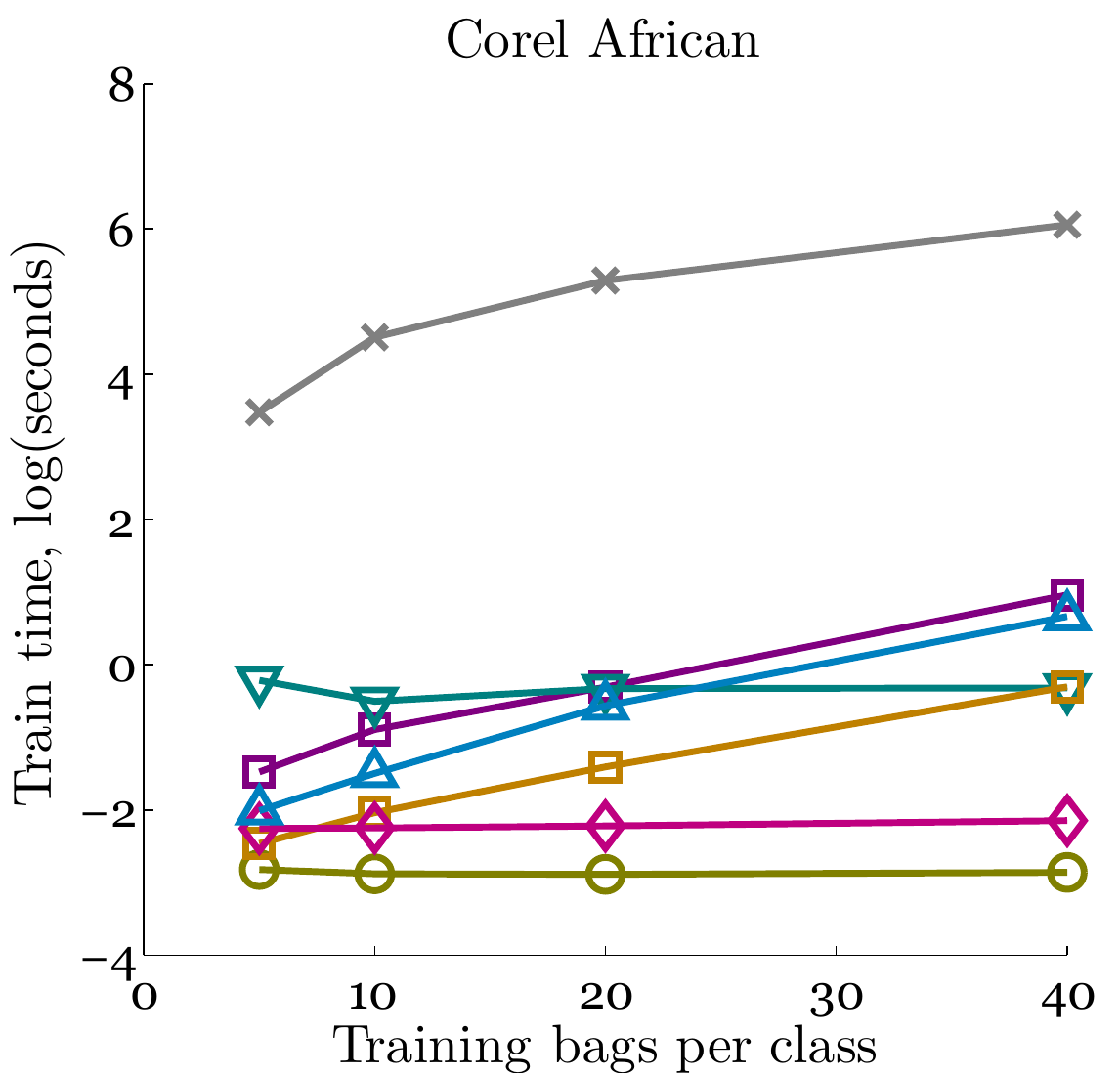}
  }

  \subfloat{
  \includegraphics[width=0.30\textwidth]{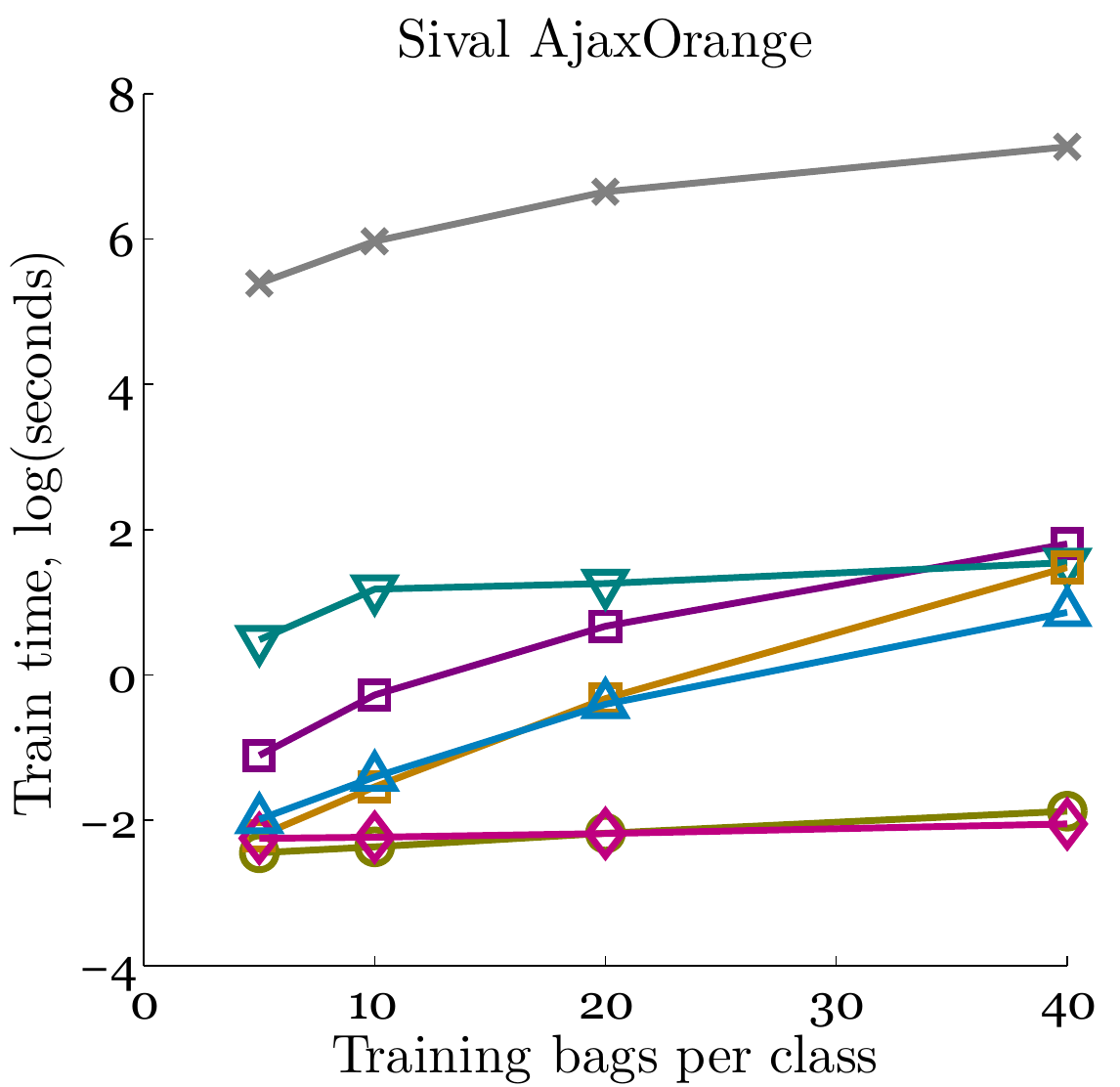}
  }
  \subfloat{
  \includegraphics[width=0.30\textwidth]{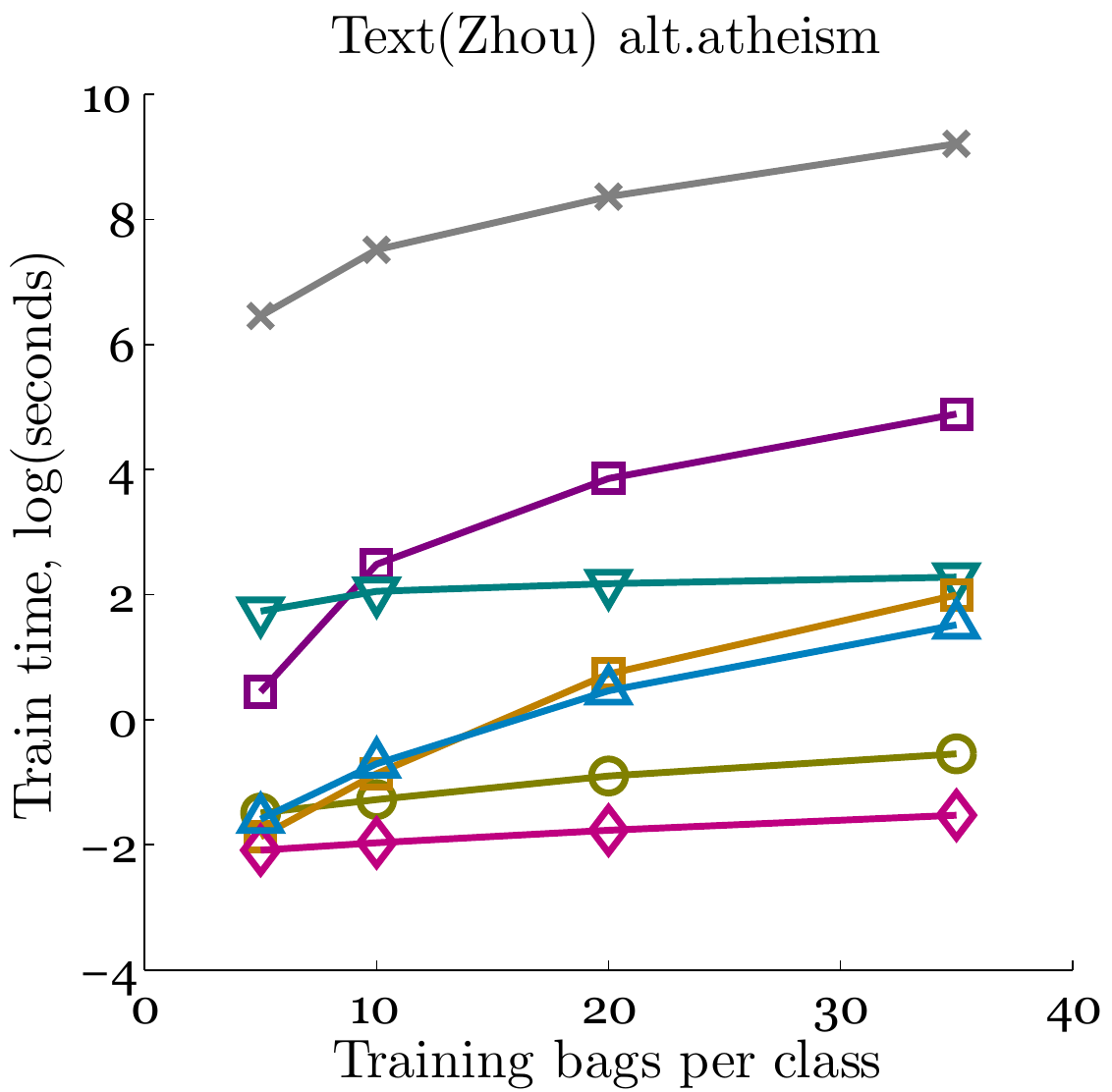}
  }
  \subfloat{
  \includegraphics[width=0.30\textwidth]{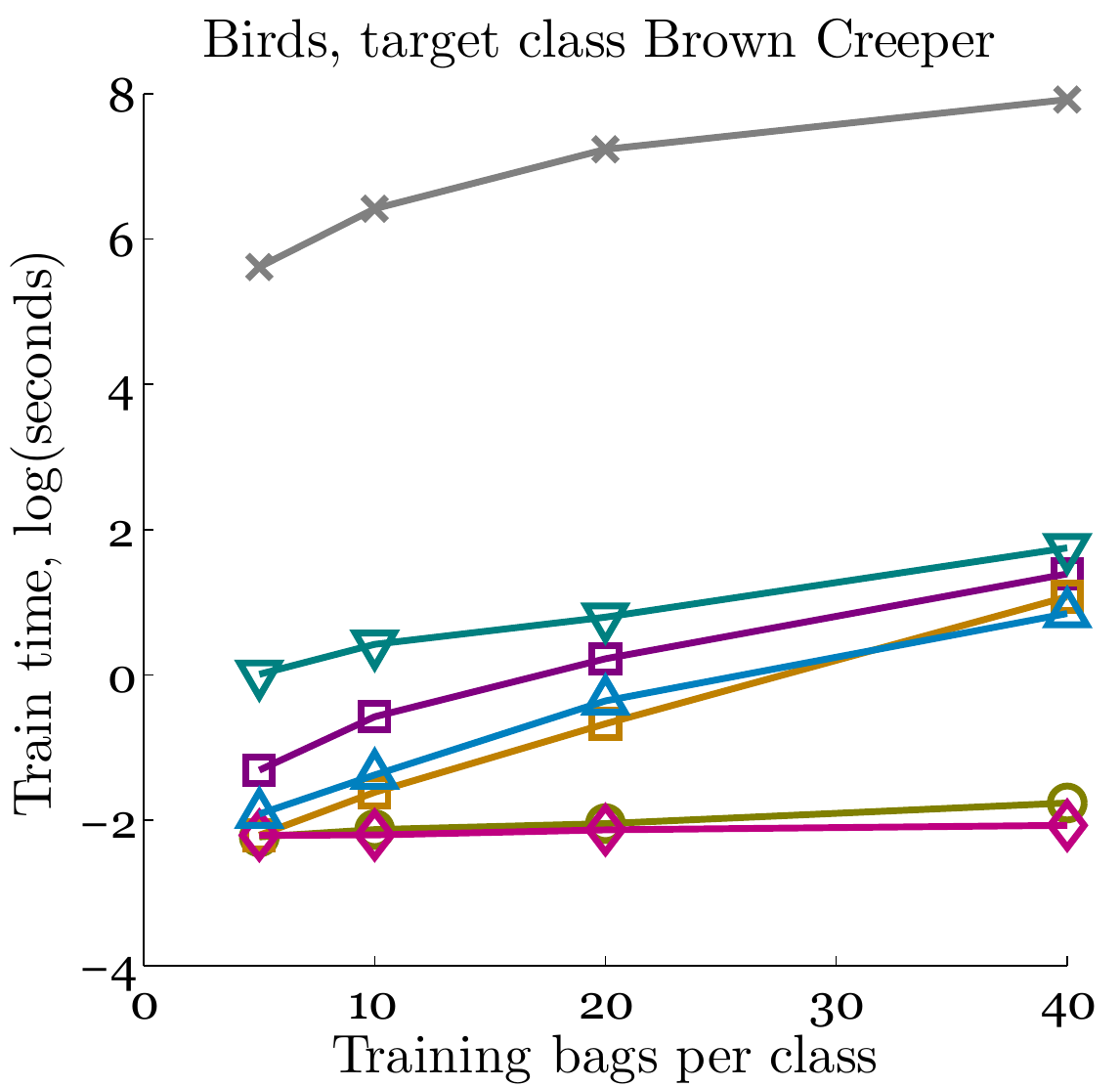}
  }
  
  \caption[]{Training time curves for Musk1, Musk2, African, AjaxOrange, alt.atheism and Brown Creeper datasets. The standard deviations in time are all quite low except for EM-DD. Figure best viewed in color.}
 \label{fig:timecurves1}
\end{figure}

%Hopefully, here we can show that:
%\begin{itemize}
%\item Learning a density (EM-DD) is very, very slow
%\item Learning a labeling may not be a well-posed problem, and is generally slower than only learing bag labels
%\item Bag-level methods perform the best and are fastest
%\end{itemize}

\subsubsection{Comparison}

As a final comparison, we present the results of 5 times 10-fold stratified cross-validation per dataset for all the available data. Unfortunately, some results cannot be reported: for EM-DD when one cross-validation fold lasts longer than five days; for MILES when there are too many instances in the training set; for MILBoost when features with the same value for all instances are present. Furthermore, for several computationally intensive methods it was infeasible to perform the experiment 50 times, therefore these performances are based on fewer runs. 

  For MILES, we use a radial basis kernel with $\sigma=10$ as the instance similarity function. MiSVM, minimax and MInD are all used with an SVM with a linear kernel and regularization parameter $C=1$.

\begin{table}[ht]
\caption{Comparison of different MIL approaches, AUC and standard error ($\times100$), $5\times10$-fold cross-validation.  \added{Significance is determined by the Friedman test, for which the critical difference is 2.0153. Classifiers in bold are best, or not significantly worse than best.}}
\centering

\resizebox{1.0\textwidth}{!}{
\begin{tabular}{l c c c c c c}
\footnotesize

& \multicolumn{6}{c}{Classifier} \\
Data & EM-DD & mi-SVM& MILBoost & MILES & Minimax & MInD \\

 \hline 
  Musk1 & 87.4 (2.1)& 81.3 (2.5)& 74.3 (2.6)&  92.8 (1.2)&  89.1 (1.9)&  93.4 (1.2)\\
    Musk2 & 86.9 (2.1)& 81.5 (2.1)& 73.6 (2.3)&  95.3 (0.8)& 89.0 (1.5)&  95.4 (1.4)\\
    Fox &  67.6 (3.2)& 53.9 (1.6)& 61.1 (1.9)&  69.8 (1.7)& 58.1 (1.3)& 60.5 (1.9)\\
    Tiger & 75.4 (2.9)&  83.3 (1.3)&  84.1 (1.6)&  87.2 (1.6)& 81.4 (1.3)&  85.1 (1.7)\\
    Elephant & 88.5 (2.1)& 84.1 (1.4)& 89.0 (1.4)& 88.3 (1.3)& 88.2 (1.0)&  93.1 (0.8)\\
    African & 91.5 (1.0)& 63.4 (1.2)& 88.9 (0.9)& 58.9 (1.7)& 84.5 (1.5)&  96.7 (0.4)\\
    Beach & 84.7 (1.3)& 49.6 (1.6)& 85.0 (1.1)& 60.0 (1.9)& 82.4 (0.9)&  92.3 (0.6)\\
    AjaxOrange & - & 93.6 (1.1)&  97.9 (0.5)& - & 91.1 (0.9)&  98.6 (0.4)\\
    Alt.atheism & 51.0 (5.2)& 70.9 (2.6)& - & 47.1 (2.4)& 80.6 (1.8)&  94.9 (1.0)\\
    Comp.graphics & 48.2 (3.2)& 59.3 (2.8)& 56.3 (2.6)& 57.2 (2.6)& 57.1 (2.7)&  92.2 (1.4)\\
    BrownCreeper & 94.5 (0.9)& 85.8 (0.7)&  95.4 (0.4)&  95.8 (0.3)& 94.1 (0.4)&  95.5 (0.3)\\
    WinterWren & 98.5 (0.3)& 95.3 (0.4)&  97.0 (1.5)&  99.2 (0.2)& 98.1 (0.2)&  99.5 (0.1)\\
    Web1 & - & 89.7 &  77.8 (5.7)&  88.2 (4.7)&  90.4 &  76.0 (2.7)\\
    Web4 & 60.6 (1.1)& 81.2 & 61.8 (4.9)& 70.8 (1.6)&  86.7 &  73.7 (3.2)\\

    \hline
    Friedman &        4.1786  &  4.3571  &  3.9286  &  \bf{3.1786}  &  \bf{3.5714}  &  {\bf 1.7857} \\

\end{tabular}
}
\label{tab:comparison}
\end{table}

%[TODO] Table needs to be updated with names. 

The results are shown in Table~\ref{tab:comparison}. \added{Overall, the best results are given by MILES and MInD. Note that MILES is sensitive to the choice of width parameter $\sigma$, and for some datasets, the default value results in poor performance. We also performed experiments with MILES with a linear kernel, and the performances were more stable across datasets, however, in general lower and significantly worse than MInD.} 
 
\added{The last row shows the results of the Friedman test~\cite{demsar2006statistical}. We have treated the missing results as random performance (AUC of 0.5) to obtain the classifier ranks. The ranks show that MInD is the best performing classifier, followed by MILES and Minimax, which are both not significantly different from MInD due to the critical difference of 2.0153 (14 datasets, 6 classifiers). MInD, however, is significantly better than the other four classifiers, which is not the case for MILES or Minimax.} 

The only methods that (i) always produce a reasonably good performance and (ii) always produce a result, are Minimax and MInD. Our advice for a different MIL problem would therefore be to try the Minimax, MInD and, if the dimensionality allows it, MILES. An additional benefit of these choices is that the costs of creating the MInD and MILES dissimilarity matrices can be shared, because both are based on instance distances.

%Hopefully, here we can show that:
%\begin{itemize}
%\item The claims based on Musk sometimes do not hold for other datasets
%\item We perform at least comparably to other methods
%\item Bag-level methods seem to be doing better overall
%\end{itemize}

\section{Recommendations}\label{sec:compare}

In this section we provide several recommendations with respect to dealing with novel MIL problems, and in particular the bag dissimilarity approach.
%In this section we outline several differences between our bag dissimilarity approach and other approaches that use a (dis)similarity-based representation. If the performances of such methods are expected to be comparable (as in many cases in the previous section), the choice of a particular method might be based on other practical issues. 

\subsection{Dissimilarity measure}

Based on our observations, $d_{meanmin}$ is a reasonable choice for many datasets. Although we can design datasets where $d_{meanmin}$ would fail, such as the Concept dataset in Fig.~\ref{fig:examples}, we have not encountered such datasets in practice. However, our advice would be to inspect 2D or 3D projections of the instances (or a subset of the instances in case of very large datasets), to see whether there are similarities between the novel dataset and any existing toy or real problems. Based on such observations, one could reason whether a particular dissimilarity function would be more or less successful. 

Another possibility is to have an expert involved in defining the dissimilarity, for instance, the underlying instance distance function could be replaced by an application-specific measure, such as an alignment measure for strings. This is likely to result in non-metric dissimilarities, which, as we have demonstrated in Section\ref{sec:props}, is not a problem for the dissimilarity approach.

For a novel MIL problem, it is necessary to consider properties of the dataset (number of bags, bag size, and the dimensionality of the instances), both in terms of performance and computational complexity. For small bags, it is difficult to estimate the instance distribution well, especially if the feature space is high-dimensional. In this case, the point set measures can be preferable. For large bags, the instance distribution can be estimated more reliably. Furthermore, the point set dissimilarities may become too expensive to compute, so a distribution measure can be preferred. In a situation where both small and large bags are present, it is important to check whether the the bag size is informative for the problem (such as a bird species singing only when a lot of birds a singing), or an artefact (such as errors in the data generation procedure).

\subsection{Classifier}

A dissimilarity representation can, in principle, be used as an input for any supervised classifier. Of course, the number of objects (bags) and the number of features (prototypes) should be taken into consideration, as in any pattern recognition problem. For instance, it might not be advisable to use a very complex classifier when the number of bags is very low. In this paper, we only performed experiments with the logistic and support vector classifiers, but in other experiments, Parzen, dissimilarity-based nearest neighbor, 1-norm SVM and others have also been successful~\cite{tax2011bag,cheplygina2012does}. This offers flexibility to a (possibly non-expert) user, who might have a preference for a certain classifier. Using sparse classifiers (such as the 1-norm SVM) can also help interpretability: the classification result can then be explained as being (dis)similar to certain prototypes.  

Another advantage of the wide choice of classifiers is that the original MIL setting (binary offline classification) can be easily transported to other learning settings. For instance, in a multi-class case with a large number of classes, it might be advantageous to use a classifier that is inherently multi-class (such as nearest neighbor) rather than combining many one-against-all binary classifiers. %Or, when it is not possible to process all the training data at once, online learning algorithms could be applied.

\section{Discussion and Conclusions}\label{sec:conclusion}

In this paper, we proposed a dissimilarity representation for multiple instance learning (MIL), where each bag is represented by its dissimilarities to the training bags. The problem is therefore converted to a supervised learning problem where any classifier can be used. There are many ways to define a dissimilarity between two bags, by viewing each bag as a point set, as a distribution in instance space, or as an attributed graph. 

We gathered a wide range of artificial and real MIL problems and discussed which are the informative instances in each case. Through experiments, we have demonstrated that different dissimilarity definitions have different implicit assumptions about the informativeness of instances, therefore making some dissimilarities more suitable than others for the dataset in question. In practice, the dissimilarity based on averaging of the minimum instance distances between bags has shown good performance in all the real-life datasets we discussed. Our approach has shown very competitive performances to other MIL algorithms, while keeping the computational effort quite low.

Furthermore, we discussed the benefits of the proposed approach to a potential end-user. Because the dissimilarity approach does not impose restrictions on the dissimilarity matrix, expert advice can be incorporated in the dissimilarity definition. Non-metric dissimilarity measures may be more informative than their metric counterparts, and such properties can be dealt with naturally in a dissimilarity approach. Furthermore, the approach is flexible with respect to the classifier used, and can be easily extended to other learning settings. 

One of the questions that is left is when to use the ``point set'' and the ``instance distribution'' approaches. Depending on the size of the bags, one of these may be more accurate than the other, however, computational issues may also become a factor. It would be interesting to investigate the exact trade-off of these two choices. %Another point we would like to investigate is how to select informative prototypes \emph{before} computing the whole dissimilarity matrix, thus reducing the amount of computation and the dimensionality of the representation. It may even be the case that prototypes do not need to be existing bags or instances, but artificially generated points or point sets. %Lastly, for problems where it is possible for a user to select the prototypes, it would be interesting to compare whether the user selection or the automatic selection is more discriminative.
Overall, we believe the proposed approach is a flexible, powerful and intuitive way to do MIL, and that combined, these qualities make it an attractive method for domains where data might be naturally grouped in bags, but MIL is not yet being used.

\section{Acknowledgements}

\added{We would like to thank the anonymous reviewers for their helpful comments on how to improve this paper.}

\bibliographystyle{elsarticle-num}	% 
\bibliography{publications}

\end{document}